\documentclass[10pt,twocolumn,letterpaper]{article}

\usepackage[pagenumbers]{cvpr} 

\usepackage{graphicx}
\usepackage{amsmath}
\usepackage{amssymb}
\usepackage{booktabs}

\usepackage{float}

\usepackage{stfloats} 

\usepackage{dsfont}
\usepackage{subcaption}
\usepackage{caption}
\usepackage{nicefrac}
\usepackage{mathrsfs}
\usepackage{xcolor}
\usepackage{enumitem}
\usepackage{colortbl}
\usepackage{multirow}

\usepackage{marvosym}
\usepackage[export]{adjustbox}

\usepackage{pifont}

\definecolor{sh_gray}{rgb}{0.84,0.84,0.84}
\definecolor{sh_gray2}{rgb}{1,0.89,0.75}
\definecolor{color3}{rgb}{0.95,0.95,0.95}
\definecolor{color4}{rgb}{0.94,0.94,1}
\definecolor{color5}{rgb}{1,0.96,0.88}

\def\xnet{Restormer\xspace}

\newlength{\Oldarrayrulewidth}

\usepackage[pagebackref=true,breaklinks,colorlinks,citecolor=blue,linkcolor=blue,urlcolor=blue]{hyperref}

\usepackage[capitalize]{cleveref}
\crefname{section}{Sec.}{Secs.}
\Crefname{section}{Section}{Sections}
\Crefname{table}{Table}{Tables}
\crefname{table}{Tab.}{Tabs.}

\def\cvprPaperID{10042} 
\def\confName{CVPR}
\def\confYear{2022}

\begin{document}


\title{\vspace{-0.5em}\xnet: Efficient Transformer for High-Resolution Image Restoration\vspace{-0.5em}}

\author{
Syed Waqas Zamir$^{1}$ \quad Aditya Arora$^{1}$  \quad Salman Khan$^{2}$ \quad Munawar Hayat$^{3}$ \\ 
Fahad Shahbaz Khan$^{2}$  \quad Ming-Hsuan Yang$^{4,5,6}$ \\
$^1$Inception Institute of AI \quad $^2$Mohamed bin Zayed University of AI \quad
$^3$Monash University\\
$^4$University of California, Merced \quad $^5$Yonsei University \quad $^6$Google Research 
\vspace{-0.5em}
}

\maketitle

\begin{abstract}\vspace{-0.5em}
Since convolutional neural networks (CNNs) perform well at learning generalizable image priors from large-scale data, these models have been extensively applied to image restoration and related tasks. Recently, another class of neural architectures, Transformers, have shown significant performance gains on natural language and high-level vision tasks. %
While the Transformer model mitigates the shortcomings of CNNs (\ie, limited receptive field and inadaptability to input content), its computational complexity grows quadratically with the spatial resolution, therefore making it infeasible to apply to most image restoration tasks involving high-resolution images. In this work, we propose an efficient Transformer model by making several key designs in the building blocks (multi-head attention and feed-forward network) such that it can capture long-range pixel interactions, while still remaining applicable to large images. Our model, named Restoration Transformer ({\xnet}), achieves state-of-the-art results on several image restoration tasks, including image deraining,  single-image motion deblurring, defocus deblurring (single-image and dual-pixel data), and image denoising (Gaussian grayscale/color denoising, and real image denoising). The source code and pre-trained models are available at \url{https://github.com/swz30/Restormer}.

\end{abstract}

\vspace{-0.5em}
\section{Introduction}
\label{sec:intro}
Image restoration is the task of reconstructing a high-quality image by removing degradations (\eg, noise, blur, rain drops)
from a degraded input. 
Due to the ill-posed nature, it is a highly challenging problem that usually requires strong image priors for effective restoration.  
Since convolutional neural networks (CNNs) perform well at learning generalizable priors from large-scale data, they have emerged as a preferable choice compared to conventional restoration approaches.

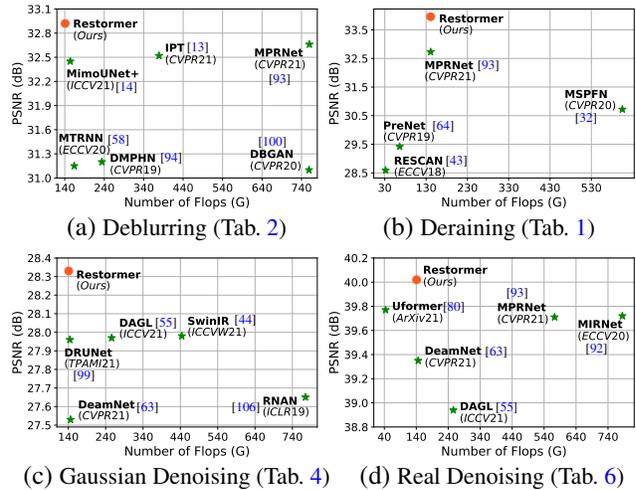
\begin{figure}[t]
  \begin{center}
    \begin{tabular}{cc}\hspace{-4mm}
    \begin{picture}(120,100)
    \put(0,0){\includegraphics[width=0.243\textwidth]{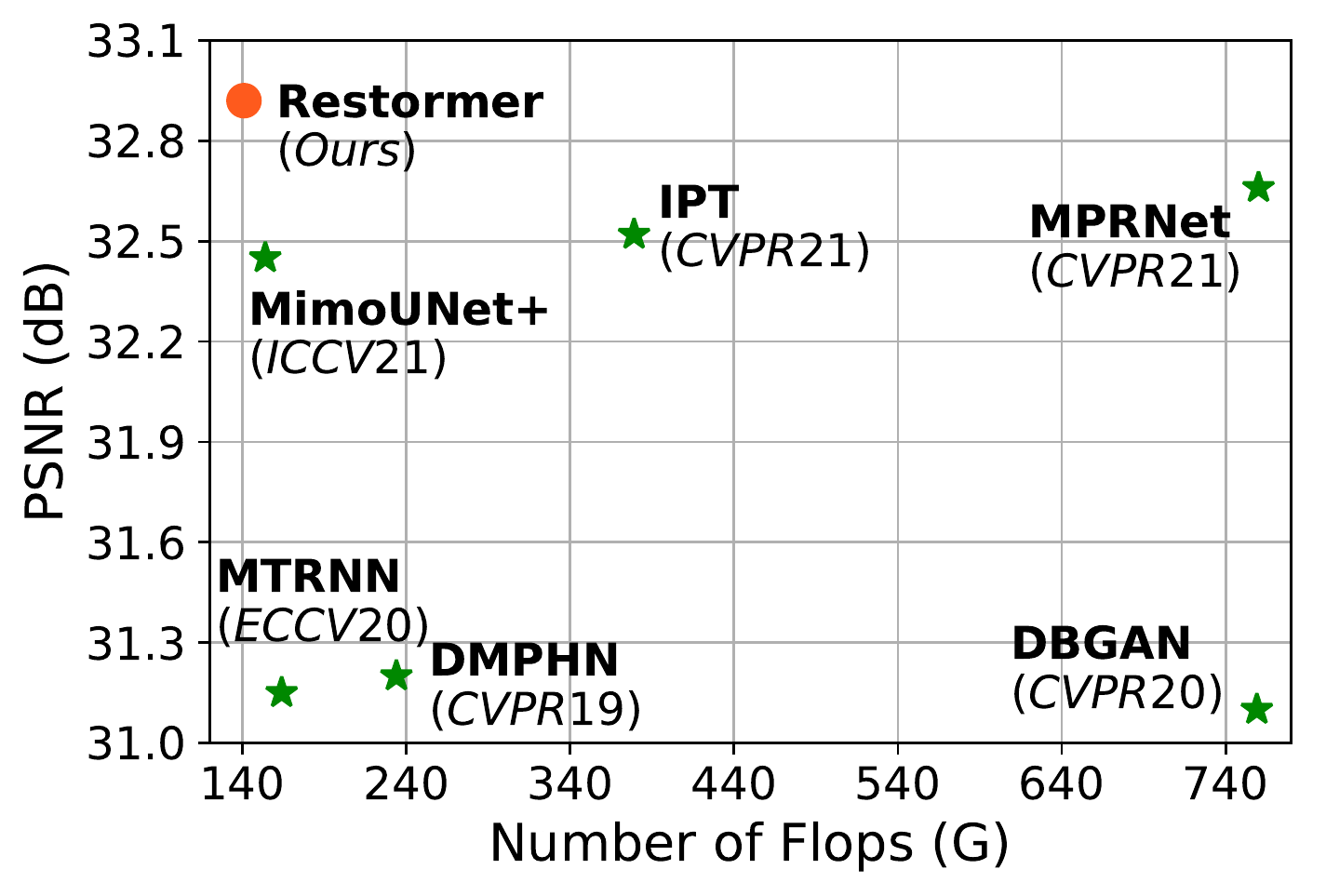}}
    \put(57,20){\tiny\cite{dmphn2019}}
    \put(37,27){\tiny\cite{mtrnn2020}}
    \put(40,47){\tiny\cite{cho2021rethinking_mimo}}
    \put(67,62){\tiny\cite{chen2021IPT}}
    \put(98,50){\tiny\cite{Zamir_2021_CVPR_mprnet}}
    \put(94,26){\tiny\cite{zhang2020dbgan}}
    \end{picture}

    & \hspace{-6mm}

    \begin{picture}(120,100)
    \put(0,0){\includegraphics[width=0.243\textwidth]{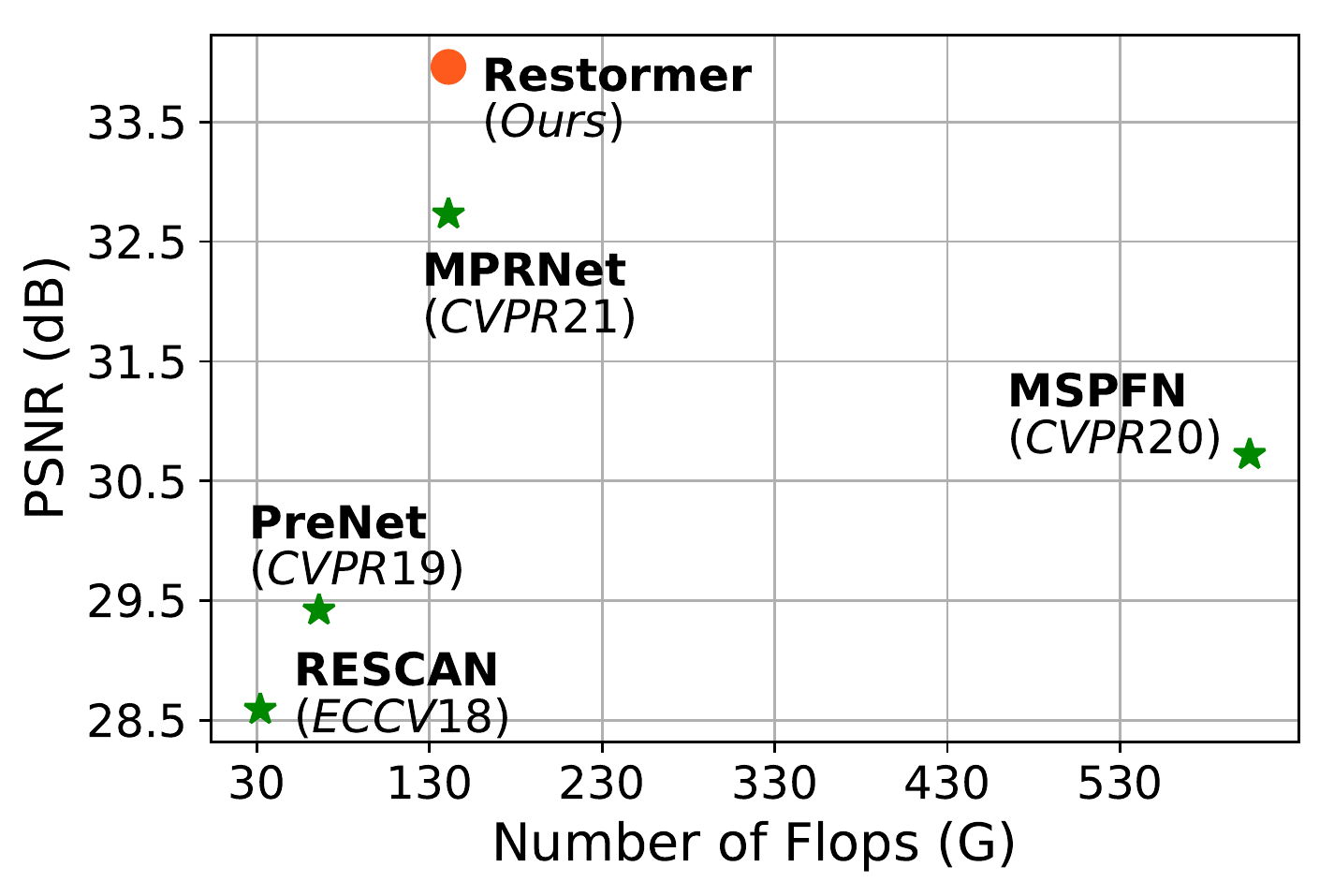}}
    \put(57,55){\tiny\cite{Zamir_2021_CVPR_mprnet}}
    \put(40,32){\tiny\cite{ren2019progressive}}
    \put(45,19){\tiny\cite{li2018recurrent}}
    \put(94,35){\tiny\cite{mspfn2020}}
    \end{picture}

    \vspace{-1.5mm}\\
    
    (a) \small Deblurring (\cref{table:deblurring}) 
    
    & \hspace{-4.2mm} 
    
    (b) \small Deraining (\cref{table:deraining}) 
    
    \vspace{-6mm} 
    \\

    \hspace{-4mm}

    \begin{picture}(120,100)
    \put(0,0){\includegraphics[width=0.243\textwidth]{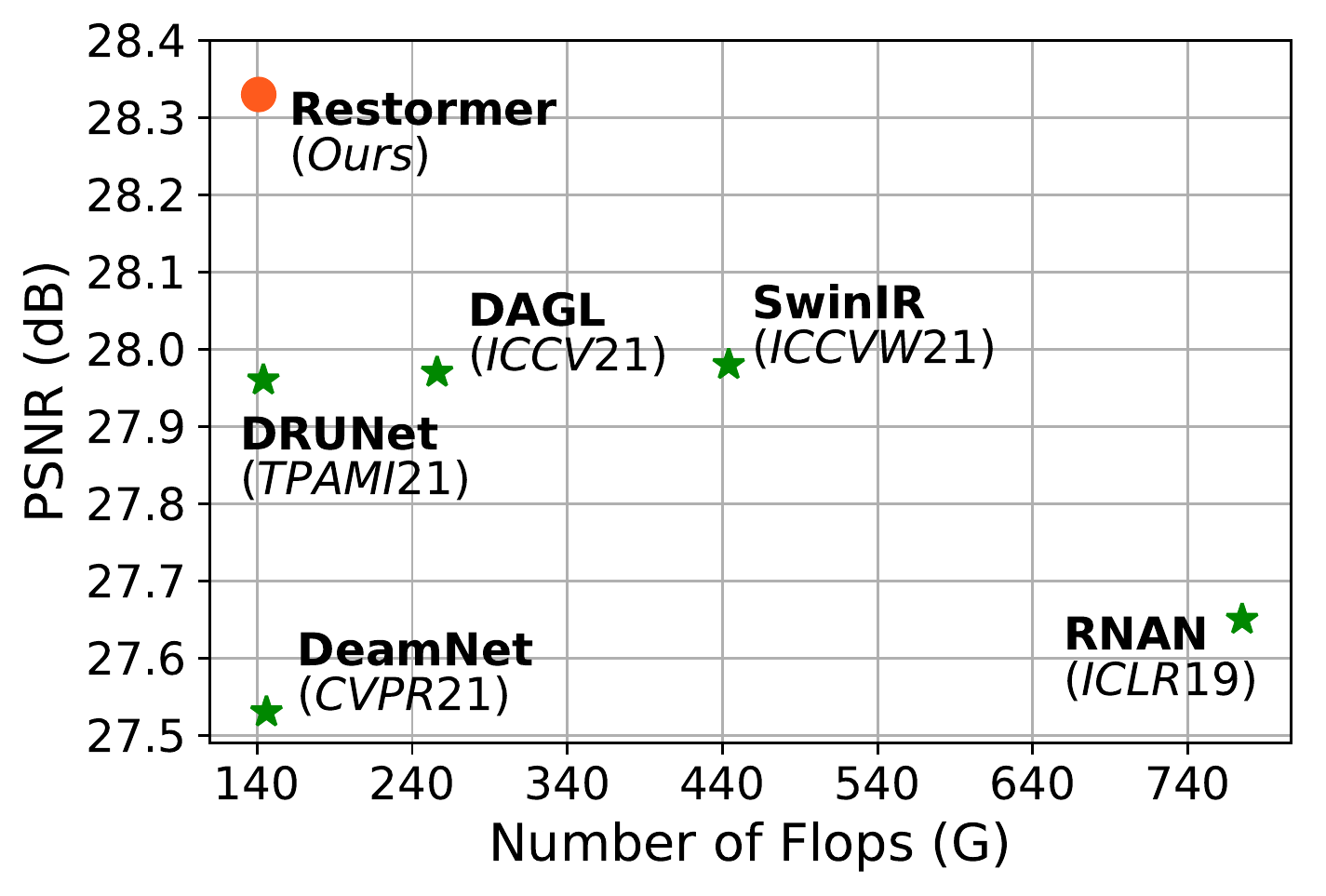}}
    \put(48,20){\tiny\cite{ren2021adaptivedeamnet}}
    \put(24,32){\tiny\cite{zhang2021DPIR}}
    \put(55,52){\tiny\cite{mou2021dynamicDAGL}}
    \put(85,53){\tiny\cite{liang2021swinir}}
    \put(84,20){\tiny\cite{zhang2019residual}}
    \end{picture}
    
    & \hspace{-6mm}

    \begin{picture}(120,100)
    \put(0,0){\includegraphics[width=0.243\textwidth]{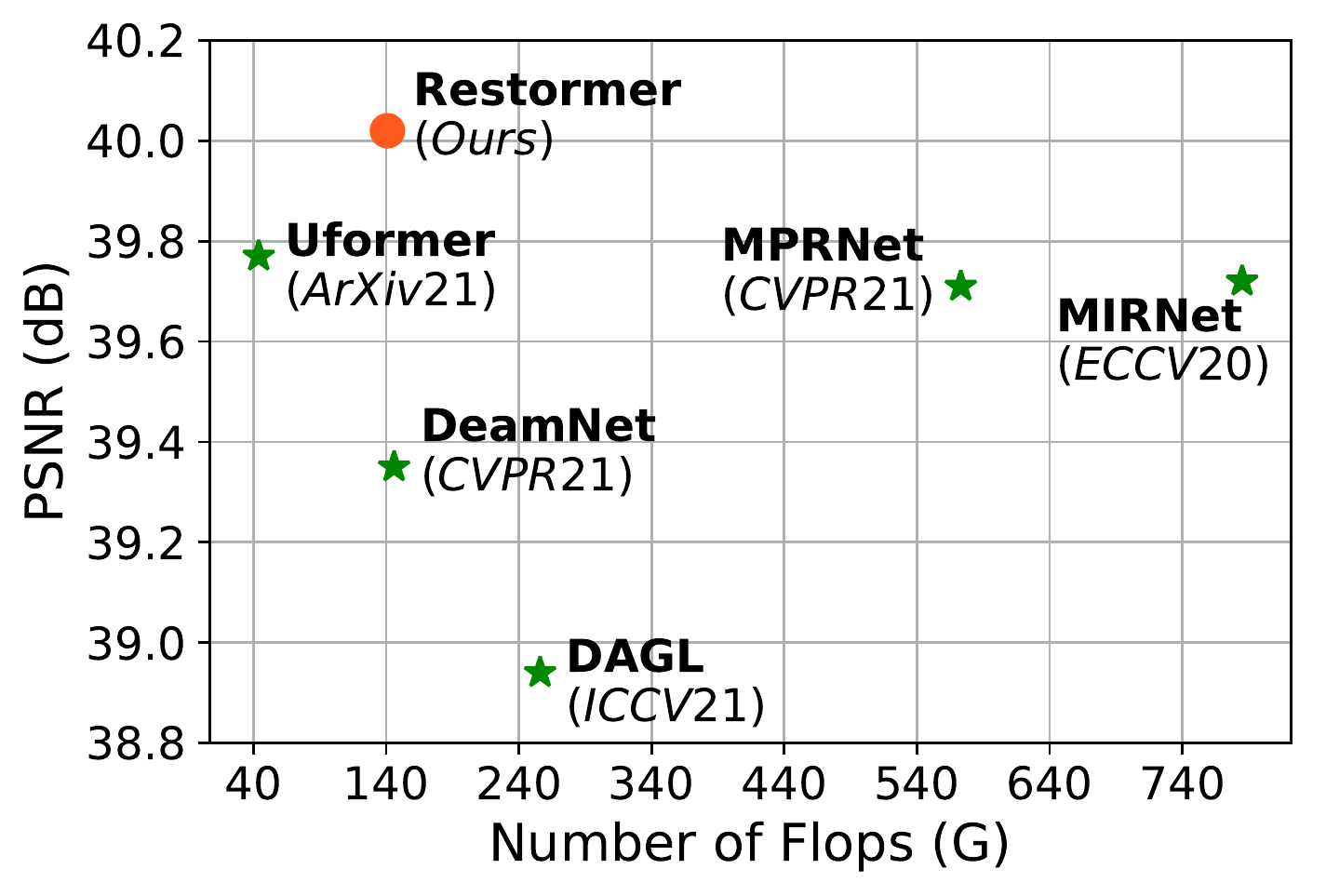}}
    \put(60,41){\tiny\cite{ren2021adaptivedeamnet}}
    \put(64,20){\tiny\cite{mou2021dynamicDAGL}}
    \put(44,58){\tiny\cite{wang2021uformer}}
    \put(68,63){\tiny\cite{Zamir_2021_CVPR_mprnet}}
    \put(98,42){\tiny\cite{zamir2020mirnet}}
    \end{picture}
    
    \vspace{-1.5mm}\\
    
    (c) \small Gaussian Denoising (\cref{table:graydenoising}) 
    
    & \hspace{-4.2mm} 
    
    (d) \small Real Denoising (\cref{table:realdenoising}) \\
    \end{tabular}
  \end{center}\vspace{-1.8em}
 \caption{Our \xnet achieves the state-of-the-art performance on image restoration tasks while being computationally efficient.}
 \vspace{-0.8em}
 \vspace{-2mm}
\label{Fig:teaser}
\end{figure}

The basic operation in CNNs is the `\emph{convolution}' that provides local connectivity and translation equivariance. 
While these properties bring efficiency and generalization to CNNs, they also cause two main issues. \textbf{(a)} The convolution operator has a limited receptive field, thus preventing it from modeling long-range pixel dependencies.
\textbf{(b)} The convolution filters have static weights at inference, and thereby cannot flexibly adapt to the input content. 
To deal with the above-mentioned shortcomings, a more powerful and dynamic alternative is the \emph{self-attention (SA)} mechanism~\cite{vaswani2017attention, wang2018non,zhang2019self_sagan,vision_transformer} that calculates response at a given pixel
by a weighted sum of all other positions. 

Self-attention is a core component in Transformer models~\cite{vaswani2017attention,khan2021transformers} but with a unique implementation, \ie, \emph{multi-head SA} that is optimized for parallelization and effective representation learning. Transformers have shown state-of-the-art performance on natural language tasks~\cite{brown2020language,liu2019roberta,radford2018improving,fedus2021switch} and on high-level vision problems~\cite{vision_transformer,touvron2021deit,carion2020end,wang2021pyramid}. 
Although SA is highly effective in capturing long-range pixel interactions, its complexity grows quadratically with the spatial resolution, therefore making it infeasible to apply to high-resolution images (a frequent case in image restoration). 
Recently, few efforts have been made to tailor Transformers for image restoration tasks~\cite{chen2021IPT,liang2021swinir,wang2021uformer}. 
To reduce the computational loads, these methods either apply SA on small spatial windows of size $8$$\times$$8$ around each pixel~\cite{liang2021swinir,wang2021uformer}, or divide {the} input image into non-overlapping patches of size $48$$\times$$48$ and compute SA on each patch independently~\cite{chen2021IPT}.  
However, restricting the spatial extent of SA is contradictory to the goal of capturing the true long-range pixel relationships, especially on high-resolution images.

In this paper, we propose an efficient Transformer for image restoration that is capable of modeling global connectivity and {is} still applicable to large images. 
Specifically, we introduce a multi-Dconv head `\emph{transposed}' attention (MDTA) block (\cref{sec:MDTA}) in place of vanilla multi-head SA~\cite{vaswani2017attention}, that has linear complexity. 
It applies SA across feature dimension rather than the spatial dimension, \ie, instead of explicitly modeling pairwise pixel interactions, MDTA computes cross-covariance across feature channels to obtain attention map from the (\emph{key} and \emph{query} projected) input features. 
An important feature of our MDTA block is the local context mixing before feature covariance computation. This is achieved via pixel-wise aggregation of cross-channel context using $1$$\times$$1$ convolution and channel-wise aggregation of local context using efficient depth-wise convolutions. This strategy provides two key advantages. 
First, it emphasizes on the spatially local context and brings in the complimentary strength of convolution operation within our pipeline. 
Second, it ensures that the contextualized global relationships between pixels are implicitly modeled while computing covariance-based attention maps.  

A feed-forward network (FN) is the other building block of the Transformer model~\cite{vaswani2017attention}, which consists of two fully connected layers with a non-linearity in between.  In this work, we reformulate the first linear transformation layer of the regular FN~\cite{vaswani2017attention} with a gating mechanism~\cite{dauphin2017language_gating} to improve the information flow through the network.
This gating layer is designed as the element-wise product of two linear projection layers, one of which is activated with the GELU non-linearity~\cite{hendrycks2016gaussian_gelu}. 
Our gated-Dconv FN (GDFN) (\cref{sec:GDFN}) is also based on local content mixing similar to {the} MDTA module to equally emphasize on the spatial context. 
The gating mechanism in GDFN controls which complementary features should flow forward and allows subsequent layers in the network hierarchy to specifically focus on more refined image attributes, thus leading to high-quality outputs. 

Apart from the above architectural novelties, we show the effectiveness of our progressive learning strategy for \xnet (\cref{{Progressive Learning}}). 
In this process, the network is trained on small patches and large batches in early epochs, and on gradually large image patches and small batches in later epochs. 
This training strategy helps \xnet to learn context from large images, and subsequently provides quality performance improvements at test time. 
We conduct comprehensive experiments and demonstrate state-of-the-art performance of our \xnet on $16$~benchmark datasets for several image restoration tasks, including image deraining, single-image motion deblurring, defocus deblurring (on single-image and dual pixel data), and image denoising (on synthetic and real data); See \cref{Fig:teaser}. 
Furthermore, we provide extensive ablations to show the effectiveness of architectural designs and experimental choices.  

\noindent The main contributions of this work are summarized below: \vspace{-1.7em}
\begin{itemize}[leftmargin=*]\setlength{\itemsep}{-0.2em}
     \item We propose \xnet, an encoder-decoder Transformer for multi-scale \emph{local-global} representation learning on high-resolution images without disintegrating them into local windows, thereby exploiting  distant image context.  
    \item We propose a multi-Dconv head transposed attention (MDTA) module that is capable of aggregating local and non-local pixel interactions, and is efficient enough to process high-resolution images.
    \item A new gated-Dconv feed-forward network (GDFN) that performs controlled feature transformation, \ie, suppressing less informative features, and allowing only the useful information to pass further through the network hierarchy. 
\end{itemize}

\begin{figure*}[t]
    \centering
    \includegraphics[width= \textwidth,valign=t]{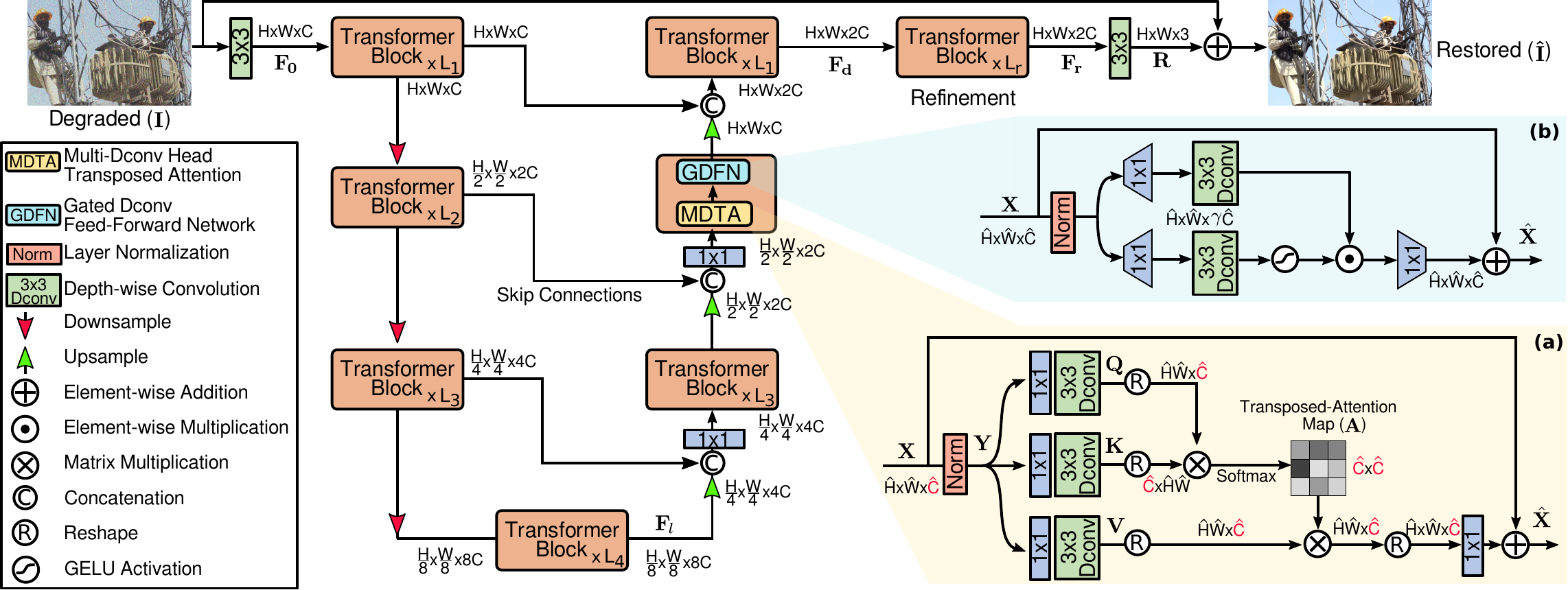}\vspace{-0.6em}
    \caption{Architecture of \xnet for high-resolution image restoration. Our \xnet consists of multiscale hierarchical design incorporating efficient Transformer blocks. The core modules of Transformer block are: \textbf{(a)} multi-Dconv head transposed attention (MDTA) that performs (spatially enriched) \emph{query}-\emph{key} feature interaction across channels rather the spatial dimension, and \textbf{(b)} Gated-Dconv feed-forward network (GDFN) that performs controlled feature transformation, \ie, to allow useful information to propagate further. }
    \label{fig:framework}
    \vspace{-1em}
\end{figure*}


\section{Background}
\label{sec:background}

\noindent \textbf{Image Restoration.} In recent years, data-driven CNN architectures \cite{rcan, zhang2020rdn, Zamir_2021_CVPR_mprnet,anwar2020densely,dudhane2021burst,zamir2020mirnet} have been shown to outperform conventional restoration approaches \cite{he2010single,timofte2013anchored,michaeli2013nonparametric,kopf2008deep}. 
Among convolutional designs, encoder-decoder based U-Net architectures \cite{deblurganv2,cho2021rethinking_mimo,Zamir_2021_CVPR_mprnet,yue2020danet,abdullah2020dpdd,wang2021uformer,zhang2021DPIR} have been predominantly studied for restoration due to their hierarchical multi-scale representation while remaining computationally efficient. 
Similarly, skip connection based approaches have been shown to be effective for restoration due to specific focus on learning residual signals \cite{zhang2019residual,liu2019dual,zamir2020mirnet,gu2019self}. 
Spatial and channel attention modules have also been incorporated to selectively attend to relevant information \cite{Zamir_2021_CVPR_mprnet,zamir2020mirnet,li2018recurrent}. 
We refer the reader to NTIRE challenge reports \cite{nah2021ntire,abuolaim2021ntire,ntire2019_denoising,ntire2019_enhancement} and recent literature reviews \cite{anwar2019deep,li2019single,tian2020deep}, which summarize major design choices for image restoration.

\noindent \textbf{Vision Transformers.} The Transformer model is first developed for sequence processing in natural language tasks \cite{vaswani2017attention}. 
It has been adapted in numerous vision tasks such as image recognition \cite{vision_transformer,touvron2021deit,yuan2021tokens}, segmentation \cite{wang2021pyramid,xie2021segformer,zheng2021rethinking}, object detection \cite{carion2020end, zhu2020deformable, liu2021swin}. 
The Vision Transformers \cite{vision_transformer, touvron2021deit} decompose an image into a sequence of patches (local windows) and learn their mutual relationships. 
The distinguishing feature of these models is the strong capability to  learn long-range dependencies between image patch sequences and adaptability to given input content \cite{khan2021transformers}. 
Due to these characteristics, Transformer models have also been studied for the low-level vision problems such as super-resolution \cite{yang2020learning, liang2021swinir}, image colorization \cite{kumar2021colorization}, denoising \cite{chen2021IPT, wang2021uformer}, and deraining \cite{wang2021uformer}. 
However, the computational complexity of SA in Transformers can increase quadratically with the number of image patches, therby prohibiting its application to high-resolution images. 
Therefore, in low-level image processing applications, where high-resolution outputs need to be generated, recent methods generally employ different strategies to reduce complexity. 
One potential remedy is to apply self-attention within local image regions \cite{liang2021swinir, wang2021uformer} using the Swin Transformer design \cite{liang2021swinir}. 
However, this design choice restricts the context aggregation within local neighbourhood, defying the main motivation of using self-attention over convolutions, thus not ideally suited for image-restoration tasks. 
In contrast, we present a Transformer model that can learn long-range dependencies while remaining 
computationally efficient.


\section{Method}\label{sec:method}
Our main goal is to develop an efficient Transformer model that can handle high-resolution images for  restoration tasks.  
To alleviate the computational bottleneck, we introduce key designs to the multi-head SA layer and a multi-scale hierarchical module that has lesser computing requirements than a single-scale network~\cite{liang2021swinir}. 
We first present the overall pipeline of our \xnet architecture (see~\cref{fig:framework}). 
Then we describe the core components of the proposed Transformer block: \textbf{(a)} multi-Dconv head transposed attention (MDTA) and \textbf{(b)} gated-Dconv feed-forward network (GDFN). 
Finally, we provide details on the progressive training scheme for effectively learning image statistics.  

\vspace{0.2em} \noindent \textbf{Overall Pipeline.} 
Given a degraded image $\mathbf{I}$~$\in$~$\mathbb{R}^{H\times W \times 3}$, \xnet first applies a convolution to obtain low-level feature embeddings $\mathbf{F_0}$~$\in$~$\mathbb{R}^{H\times W \times C}$; where $H\times W$ denotes the spatial dimension and $C$ is the number of channels. Next, these shallow features $\mathbf{F_0}$ pass through a 4-level symmetric encoder-decoder and transformed into deep features $\mathbf{F_d}$~$\in$~$\mathbb{R}^{H\times W \times 2C}$. 
Each level of encoder-decoder contains multiple Transformer blocks, where the number of blocks are gradually increased from the top to bottom levels to maintain efficiency. 
Starting from the high-resolution input, the encoder hierarchically reduces spatial size, while expanding channel capacity. The decoder takes low-resolution latent features $\mathbf{F}_l$~$\in$~$\mathbb{R}^{\frac{H}{8}\times\frac{W}{8}\times 8C}$ as input and progressively recovers the high-resolution representations. 
For feature downsampling and upsampling, we apply pixel-unshuffle and pixel-shuffle operations~\cite{shi2016real_pixelshuffle}, respectively. 
To assist {the} recovery process, the encoder features are concatenated with the decoder features via skip connections~\cite{ronneberger2015unet}. 
The concatenation operation is followed by a $1$$\times$$1$ convolution to reduce channels (by half) at all levels, except the top one. 
At level-1, we let Transformer blocks to aggregate the low-level image features of the encoder with the high-level features of the decoder. It is beneficial in preserving the fine structural and textural details in the restored images.
Next, the deep features $\mathbf{F_d}$ are further enriched in the refinement stage operating at high spatial resolution.  These design choices yield quality improvements as we shall see in the experiment section (\cref{sec:experiments}). 
Finally, a convolution layer is applied to the refined features to generate residual image $\mathbf{R}$~$\in$~$\mathbb{R}^{H\times W \times 3}$ to which degraded image is added to obtain the restored image: $\mathbf{\hat{I}} = \mathbf{I} + \mathbf{R}$. Next, we present the modules of the Transformer block. 

\subsection{Multi-Dconv Head Transposed Attention}\label{sec:MDTA}
The major computational overhead in Transformers comes from the self-attention layer. In conventional SA~\cite{vaswani2017attention,vision_transformer}, the time and memory complexity of the key-query dot-product interaction grows quadratically with the spatial resolution of input, \ie, $\mathcal{O}(W^2H^2)$ for images of $W$$\times$$H$ pixels. 
Therefore, it is infeasible to apply SA on most image restoration tasks that often involve high-resolution images. 
To alleviate this issue, we propose MDTA, shown in~\cref{fig:framework}\textcolor{blue}{(a)}, that has linear complexity. 
The key ingredient is to apply SA across channels rather than the spatial dimension, \ie, to compute cross-covariance across channels to generate {an} attention map encoding the global context implicitly. 
As another essential component in MDTA, we introduce {depth-wise convolutions} to emphasize on {the} local context before computing feature covariance to produce the global attention map. 
From a layer normalized tensor $\mathbf{Y}$~$\in$~$\mathbb{R}^{\hat{H}\times \hat{W} \times \hat{C}}$, our MDTA first generates \emph{query} (\textbf{Q}), \emph{key} (\textbf{K}) and \emph{value} (\textbf{V}) projections, enriched with local context. It is achieved by applying $1$$\times$$1$ convolutions to aggregate pixel-wise cross-channel context followed by $3$$\times$$3$ depth-wise convolutions to encode channel-wise spatial context, yielding $\textbf{Q}{=}W_d^QW_p^Q{\textbf{Y}}$, $\textbf{K}{=}W_d^KW_p^K{\textbf{Y}}$ and $\textbf{V}{=}W_d^VW_p^V{\textbf{Y}}$.  Where $W^{(\cdot)}_p$ is the $1$$\times$$1$ point-wise convolution and $W^{(\cdot)}_d$ is the $3$$\times$$3$ depth-wise convolution. We use bias-free convolutional layers in the network. Next, we reshape query and key projections such that their dot-product interaction generates a transposed-attention map $\mathbf{A}$ of size $\mathbb{R}^{\hat{C}\times \hat{C}}$, instead of the huge regular attention map of size $\mathbb{R}^{\hat{H}\hat{W}\times \hat{H}\hat{W}}$ \cite{vision_transformer,vaswani2017attention}. Overall, the MDTA process  is defined as:
\begin{equation}
    \begin{split}
        &\textrm{$\hat{\textbf{X}}$} = W_p\,\textrm{Attention}\left(\hat{\textbf{Q}}, \hat{\textbf{K}}, \hat{\textbf{V}}\right) + \textbf{X}, 
        \\
        &\textrm{Attention}\left(\hat{\textbf{Q}}, \hat{\textbf{K}}, \hat{\textbf{V}}\right) = \hat{\textbf{V}} \cdot \textrm{Softmax$\left( \hat{\textbf{K}} \cdot \hat{\textbf{Q}}/\alpha \right)$},  
    \end{split}
\end{equation}
where $\mathbf{X}$ and $\hat{\mathbf{X}}$ are the input and output feature maps; $\hat{\mathbf{Q}} \in \mathbb{R}^{\hat{H}\hat{W}\times \hat{C}}$;  $\hat{\mathbf{K}} \in \mathbb{R}^{ \hat{C}\times\hat{H}\hat{W}}$; and $\hat{\mathbf{V}} \in \mathbb{R}^{\hat{H}\hat{W}\times \hat{C}}$ matrices are obtained after reshaping tensors from the original size $\mathbb{R}^{\hat{H}\times \hat{W} \times \hat{C}}$. 
Here, $\alpha$ is a learnable scaling parameter to control the magnitude of the dot product of $\hat{\mathbf{K}}$ and $\hat{\mathbf{Q}}$ before applying the softmax function. Similar to the conventional multi-head SA~\cite{vision_transformer}, we divide the number of channels into `heads' and learn separate attention maps {in parallel}.

\subsection{Gated-Dconv Feed-Forward Network}\label{sec:GDFN}
To transform features, the regular feed-forward network (FN)~\cite{vaswani2017attention,vision_transformer} operates on each pixel location separately and identically. It uses two $1$$\times$$1$ convolutions, one to expand the feature channels (usually by factor $\gamma$$=$$4$) and second to reduce channels back to the original input dimension. A non-linearity is applied in the hidden layer. 
In this work, we propose two fundamental modifications in FN to improve representation learning: (1) gating mechanism, and (2) depthwise convolutions. The architecture of our GDFN is shown in~\cref{fig:framework}\textcolor{blue}{(b)}. The gating mechanism is formulated as the element-wise product of two parallel paths of linear transformation layers, one of which is activated with the GELU non-linearity~\cite{hendrycks2016gaussian_gelu}. 
As in MDTA, we also include depth-wise convolutions in GDFN to encode information from spatially neighboring pixel positions, useful for learning local image structure for effective restoration. Given an input tensor $\mathbf{X}$~$\in$~$\mathbb{R}^{\hat{H}\times \hat{W} \times \hat{C}}$,  GDFN is formulated as:\vspace{-0.5em}
\begin{equation}
    \begin{split}
        \textrm{$\hat{\textbf{X}}$} &= W^0_p\,\textrm{Gating}\left(\textbf{X}\right) + \textbf{X}, 
        \\
        \textrm{Gating}(\textbf{X}) &= {\phi}(  W^1_d W^1_p(\textrm{LN}(\textbf{X})) ) {\odot} W^2_d W^2_p(\textrm{LN}(\textbf{X})), 
    \end{split}\vspace{-0.5em}
\end{equation}
where $\odot$ denotes element-wise multiplication, $\phi$ represents the GELU non-linearity, and LN is the layer normalization~\cite{ba2016layer}. {Overall, the GDFN controls the information flow through the respective hierarchical levels in our pipeline, thereby allowing each level to focus on the fine details complimentary to the other levels.} 
That is, GDFN offers a distinct role compared to MDTA (focused on enriching features with contextual information). 
Since the proposed GDFN performs more operations as compared to the regular FN~\cite{vision_transformer}, we reduce the expansion ratio $\gamma$ so {as} to have similar parameters and compute burden.


\begin{table*}
\begin{center}
\caption{\small \underline{\textbf{Image deraining}} results. When averaged across all five datasets, our \xnet advances state-of-the-art by $1.05$~dB.}
\label{table:deraining}
\vspace{-2mm}
\setlength{\tabcolsep}{9.5pt}
\scalebox{0.7}{
\begin{tabular}{l c c c c c c c c c c || c c}
\toprule[0.15em]
  & \multicolumn{2}{c}{\textbf{Test100}~\cite{zhang2019image}}&\multicolumn{2}{c}{\textbf{Rain100H}~\cite{yang2017deep}}&\multicolumn{2}{c}{\textbf{Rain100L}~\cite{yang2017deep}}&\multicolumn{2}{c}{\textbf{Test2800}~\cite{fu2017removing}}&\multicolumn{2}{c||}{\textbf{Test1200}~\cite{zhang2018density}}&\multicolumn{2}{c}{\textbf{Average}}\\
 \textbf{Method} & PSNR~$\textcolor{black}{\uparrow}$ & SSIM~$\textcolor{black}{\uparrow}$ & PSNR~$\textcolor{black}{\uparrow}$ & SSIM~$\textcolor{black}{\uparrow}$ & PSNR~$\textcolor{black}{\uparrow}$ & SSIM~$\textcolor{black}{\uparrow}$ & PSNR~$\textcolor{black}{\uparrow}$ & SSIM~$\textcolor{black}{\uparrow}$ & PSNR~$\textcolor{black}{\uparrow}$ & SSIM~$\textcolor{black}{\uparrow}$ & PSNR~$\textcolor{black}{\uparrow}$ & SSIM~$\textcolor{black}{\uparrow}$\\
\midrule[0.15em]
DerainNet~\cite{fu2017clearing} & 22.77  & 0.810  & 14.92  & 0.592  & 27.03  & 0.884  & 24.31  & 0.861  & 23.38  & 0.835  & 22.48 & 0.796    \\
SEMI~\cite{wei2019semi} & 22.35&0.788& 16.56&0.486& 25.03&0.842& 24.43&0.782& 26.05&0.822 & 22.88 & 0.744 \\
DIDMDN~\cite{zhang2018density} & 22.56&0.818& 17.35&0.524& 25.23&0.741& 28.13&0.867& 29.65&0.901& 24.58 & 0.770 \\
UMRL~\cite{yasarla2019uncertainty} & 24.41&0.829& 26.01&0.832& 29.18&0.923& 29.97&0.905& 30.55&0.910& 28.02 & 0.880 \\
RESCAN~\cite{li2018recurrent} & 25.00&0.835& 26.36&0.786& 29.80&0.881& 31.29&0.904& 30.51&0.882& 28.59 & 0.857 \\
PreNet~\cite{ren2019progressive} & 24.81&0.851& 26.77&0.858& 32.44 & 0.950 & 31.75&0.916& 31.36&0.911& 29.42  & 0.897 \\
MSPFN~\cite{mspfn2020}  & 27.50 & 0.876 & 28.66 & 0.860 & 32.40 & 0.933 & 32.82 & 0.930 & 32.39 & 0.916 & 30.75 & 0.903\\
MPRNet~\cite{Zamir_2021_CVPR_mprnet}  & 30.27 & 0.897 & 30.41 & 0.890 & 36.40 & 0.965 & \underline{33.64} & \underline{0.938} & 32.91 & 0.916 & 32.73 & 0.921 \\
SPAIR~\cite{purohit2021spatially_spair} & \underline{30.35} & \underline{0.909} & \underline{30.95} & \underline{0.892} & \underline{36.93} & \underline{0.969} & 33.34 & 0.936 & \underline{33.04} &\underline{0.922} & \underline{32.91} & \underline{0.926} \\
\midrule
\textbf{\xnet} & \textbf{32.00} & \textbf{0.923} & \textbf{31.46} & \textbf{0.904} & \textbf{38.99} & \textbf{0.978} & \textbf{34.18} & \textbf{0.944} & \textbf{33.19} & \textbf{0.926} & \textbf{33.96} & \textbf{0.935} \\
\bottomrule[0.1em]
\end{tabular}}
\end{center}\vspace{-1.5em}
\end{table*}

\begin{figure*}[!t]
\begin{center}
\scalebox{0.97}{
\begin{tabular}[b]{c@{ } c@{ }  c@{ } c@{ } c@{ } c@{ }	}\hspace{-4mm}
    \multirow{4}{*}{\includegraphics[width=.326\textwidth,valign=t]{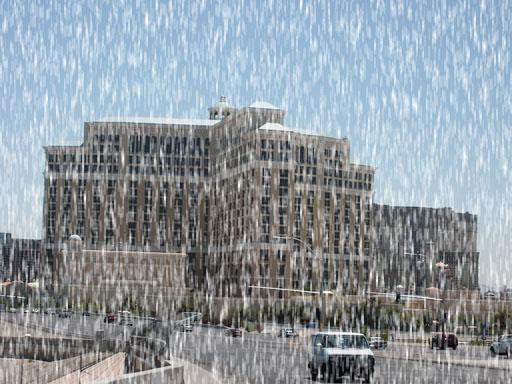}} &   
    \includegraphics[trim={ 192 87 250 244
 },clip,width=.13\textwidth,valign=t]{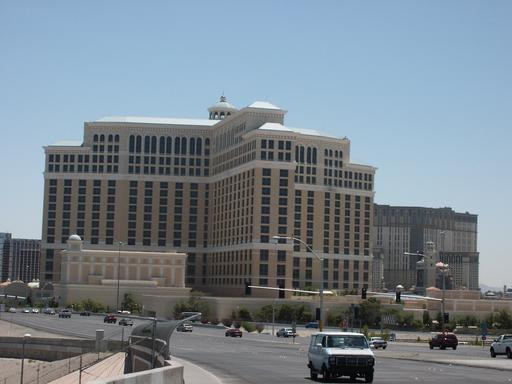}&
  	\includegraphics[trim={192 87 250 244 },clip,width=.13\textwidth,valign=t]{Images/Deraining/input_18_76.jpg}&   
    \includegraphics[trim={192 87 250 244 },clip,width=.13\textwidth,valign=t]{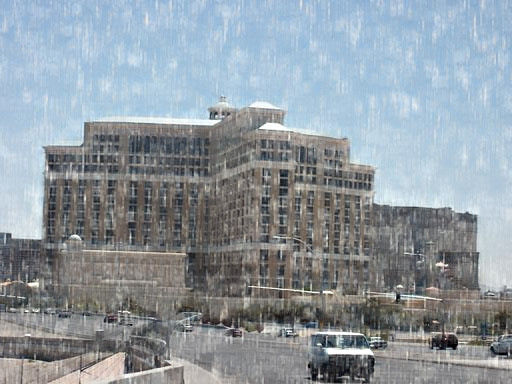}&
      \includegraphics[trim={192 87 250 244 },clip,width=.13\textwidth,valign=t]{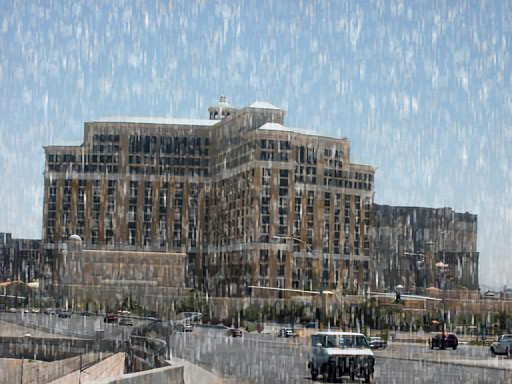}&
      \includegraphics[trim={192 87 250 244 },clip,width=.13\textwidth,valign=t]{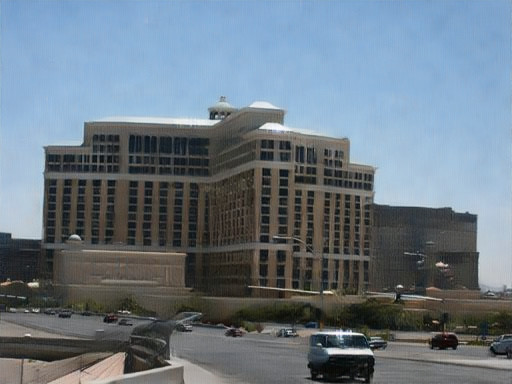}
\\
    &  \small~PSNR &\small~18.76 dB  & \small~20.23 dB &  \small~23.66 dB & \small~25.52 dB  \\
    & \small~Reference & \small~Rainy  & \small~DerainNet~\cite{fu2017clearing}  & \small~SEMI~\cite{wei2019semi}& \small~UMRL~\cite{yasarla2019uncertainty} \\
    & \includegraphics[trim={192 87 250 244 },clip,width=.13\textwidth,valign=t]{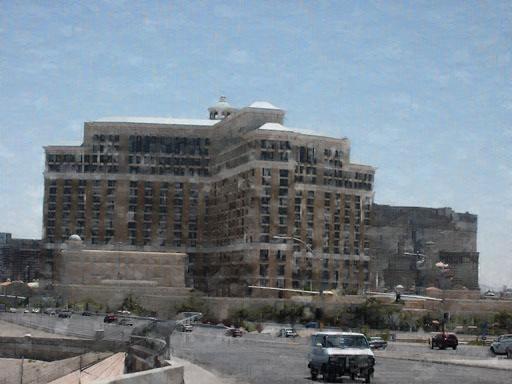}&
    \includegraphics[trim={192 87 250 244 },clip,width=.13\textwidth,valign=t]{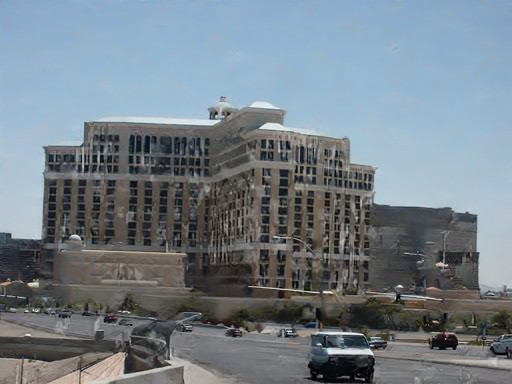}&  
     \includegraphics[trim={192 87 250 244 },clip,width=.13\textwidth,valign=t]{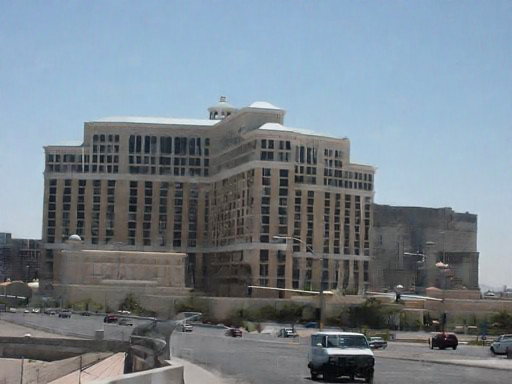}&
     \includegraphics[trim={192 87 250 244 },clip,width=.13\textwidth,valign=t]{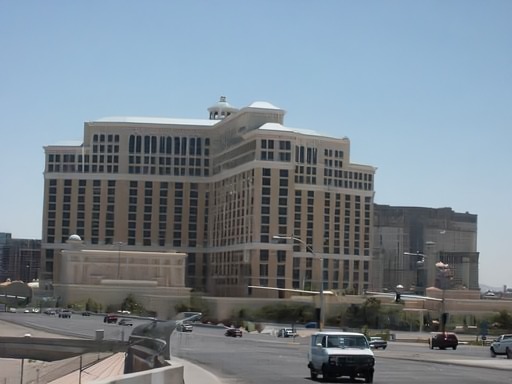}&
        \includegraphics[trim={192 87 250 244 },clip,width=.13\textwidth,valign=t]{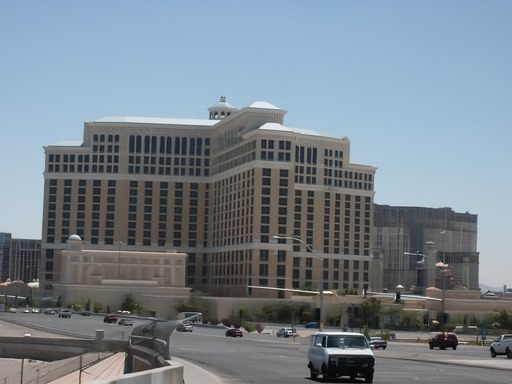}\\
     \small~18.76 dB& \small~26.88 dB & \small~27.16 dB 
     & \small~29.86 dB & 32.15 dB& \small~\textbf{33.97 dB}\\
           \small~Rainy Image  & \small~RESCAN~\cite{li2018recurrent} & \small~PreNet~\cite{ren2019progressive} & \small~MSPFN~\cite{mspfn2020}   & \small MPRNet~\cite{Zamir_2021_CVPR_mprnet} & \small~\textbf{\xnet}
\hspace{-2mm}
\end{tabular}}
\end{center}
\vspace*{-6mm}
\caption{\small \underline{\textbf{Image deraining}} example. Our \xnet generates rain-free image with structural fidelity and without artifacts. 
}
\label{fig:deraining}
\vspace{-1em}
\end{figure*}


\subsection{Progressive Learning}\label{Progressive Learning}
CNN-based restoration models are usually trained on fixed-size image patches. 
However, training a Transformer model on small cropped patches may not encode the global image statistics, thereby providing suboptimal performance on full-resolution images at test time.  To this end, we perform progressive learning where the network is trained on smaller image patches in the early epochs and on gradually larger patches in the later training epochs. The model trained on mixed-size patches via progressive learning shows enhanced performance at test time where images can be of different resolutions (a common case in image restoration). The progressive learning strategy behaves in a similar fashion to the curriculum learning process where the network starts with a simpler task and gradually moves to learning a more complex one (where the preservation of fine image structure/textures is required). Since training on large patches comes at the cost of longer time, we reduce the batch size as the patch size increases to maintain {a} similar time per optimization step as of the fixed patch training.



\begin{table}[!t]
\begin{center}
\caption{\small \underline{\textbf{Single-image motion deblurring}} results. Our \xnet is trained only on the GoPro dataset~\cite{gopro2017} and directly applied to the HIDE~\cite{shen2019human} and RealBlur~\cite{rim_2020_realblur} benchmark datasets.}
\label{table:deblurring}
\vspace{-3mm}
\setlength{\tabcolsep}{1.9pt}
\scalebox{0.69}{
\begin{tabular}{l c | c | c | c }
\toprule[0.15em]
 & \textbf{GoPro}~\cite{gopro2017} & \textbf{HIDE}~\cite{shen2019human} & \textbf{RealBlur-R}~\cite{rim_2020_realblur} & \textbf{\textbf{RealBlur-J}}~\cite{rim_2020_realblur} \\
 \textbf{Method} & PSNR~\colorbox{color4}{SSIM} & PSNR~\colorbox{color4}{SSIM} & PSNR~\colorbox{color4}{SSIM} & PSNR~\colorbox{color4}{SSIM}\\
\midrule[0.15em]
Xu \etal \cite{xu2013unnatural}     & 21.00 \colorbox{color4}{0.741} & -  &   34.46 \colorbox{color4}{0.937} &  27.14 \colorbox{color4}{0.830} \\
DeblurGAN \cite{deblurgan}          & 28.70 \colorbox{color4}{0.858} & 24.51 \colorbox{color4}{0.871} & 33.79 \colorbox{color4}{0.903}  &  27.97 \colorbox{color4}{0.834} \\
Nah \etal \cite{gopro2017}          & 29.08 \colorbox{color4}{0.914} & 25.73 \colorbox{color4}{0.874}  &  32.51 \colorbox{color4}{0.841}  &  27.87 \colorbox{color4}{0.827} \\
Zhang \etal \cite{zhang2018dynamic} & 29.19 \colorbox{color4}{0.931} & - &  35.48 \colorbox{color4}{0.947}  &  27.80 \colorbox{color4}{0.847}\\
\small{DeblurGAN-v2 \cite{deblurganv2}}    & 29.55 \colorbox{color4}{0.934} & 26.61 \colorbox{color4}{0.875} & 35.26 \colorbox{color4}{0.944}  &  28.70 \colorbox{color4}{0.866} \\
SRN~\cite{tao2018scale}             & 30.26 \colorbox{color4}{0.934} & 28.36 \colorbox{color4}{0.915} & 35.66 \colorbox{color4}{0.947} &  28.56  \colorbox{color4}{0.867} \\
Shen \etal \cite{shen2019human}     & -                         & 28.89 \colorbox{color4}{0.930} & -        & -  \\
Gao \etal \cite{gao2019dynamic}     & 30.90 \colorbox{color4}{0.935} &  29.11 \colorbox{color4}{0.913}  & -      & - \\
DBGAN \cite{zhang2020dbgan}         & 31.10 \colorbox{color4}{0.942} & 28.94 \colorbox{color4}{0.915}  & 33.78 \colorbox{color4}{0.909}     & 24.93 \colorbox{color4}{0.745} \\
MT-RNN \cite{mtrnn2020}             & 31.15 \colorbox{color4}{0.945} & 29.15 \colorbox{color4}{0.918}   & 35.79 \colorbox{color4}{0.951}     & 28.44 \colorbox{color4}{0.862}\\
DMPHN \cite{dmphn2019}              & 31.20 \colorbox{color4}{0.940}  & 29.09 \colorbox{color4}{0.924} &  35.70 \colorbox{color4}{0.948} & 28.42 \colorbox{color4}{0.860} \\
Suin \etal \cite{Maitreya2020}      & 31.85 \colorbox{color4}{0.948} & 29.98 \colorbox{color4}{0.930} & -       & - \\
SPAIR~\cite{purohit2021spatially_spair} & 32.06 \colorbox{color4}{0.953} & 30.29 \colorbox{color4}{0.931} & - & \underline{28.81} \colorbox{color4}{\underline{0.875}}\\
MIMO-UNet+~\cite{cho2021rethinking_mimo} & 32.45 \colorbox{color4}{0.957} & 29.99 \colorbox{color4}{0.930} & 35.54 \colorbox{color4}{0.947} & 27.63 \colorbox{color4}{0.837}\\
IPT~\cite{chen2021IPT} & \hspace{-2em} 32.52 \colorbox{color4}{-} & - & - & -\\
MPRNet~\cite{Zamir_2021_CVPR_mprnet} & \underline{32.66} \colorbox{color4}{\underline{0.959}} &	\underline{30.96} \colorbox{color4}{\underline{0.939}} & \underline{35.99}   \colorbox{color4}{\underline{0.952}} & 28.70 \colorbox{color4}{0.873} \\
\bottomrule[0.1em]
\textbf{\xnet} & \textbf{32.92} \colorbox{color4}{\textbf{0.961}} & \textbf{31.22} \colorbox{color4}{\textbf{0.942}} & \textbf{36.19} \colorbox{color4}{\textbf{0.957}} & \textbf{28.96} \colorbox{color4}{\textbf{0.879}}\\
\bottomrule[0.1em]
\end{tabular}}
\end{center}\vspace{-2em}
\end{table}

\section{Experiments and Analysis}\label{sec:experiments}
We evaluate the proposed \xnet on benchmark datasets and experimental settings for four image processing tasks: \textbf{(a)} image deraining, \textbf{(b)} single-image motion deblurring, \textbf{(c)} defocus deblurring (on single-image, and dual-pixel data), and \textbf{(d)} image denoising (on synthetic and real data). 
More details on datasets, training protocols, and additional visual results are presented in the supplementary material. 
In tables, the best and second-best quality scores of the evaluated methods are \textbf{highlighted} and \underline{underlined}. 

\noindent\textbf{Implementation Details.}
We train separate models for different image restoration tasks. 
In all experiments, we use the following training parameters, unless mentioned otherwise. Our \xnet employs a 4-level encoder-decoder.
From level-1 to level-4, the number of Transformer blocks are [$4,6,6,8$], attention heads in MDTA are [$1,2,4,8$], and number of channels are $[48,96,192,384]$. 
The refinement stage contains $4$ blocks. The channel expansion factor in GDFN is $\gamma{=}2.66$. 
We train models with AdamW optimizer ($\beta_1{=}0.9$, $\beta_2{=}0.999$, weight decay $1e^{-4}$) and L$_1$ loss for $300$K iterations with the initial learning rate $3e^{-4}$  gradually reduced to $1e^{-6}$ with the cosine annealing~\cite{loshchilov2016sgdr}.
For progressive learning, we start training with patch size $128$$\times$$128$ and batch size $64$. 
The patch size and batch size pairs are updated to [($160^2$,$40$), ($192^2$,$32$), ($256^2$,$16$), ($320^2$,$8$), ($384^2$,$8$)] at iterations [$92$K, $156$K, $204$K, $240$K, $276$K]. For data augmentation, we use horizontal and vertical flips.


\begin{figure*}[!t]
\begin{center}
\scalebox{0.97}{
\begin{tabular}[b]{c@{ } c@{ }  c@{ } c@{ } c@{ } c@{ }	}
\hspace{-4mm}
    \multirow{4}{*}{\includegraphics[width=.3\textwidth,valign=t]{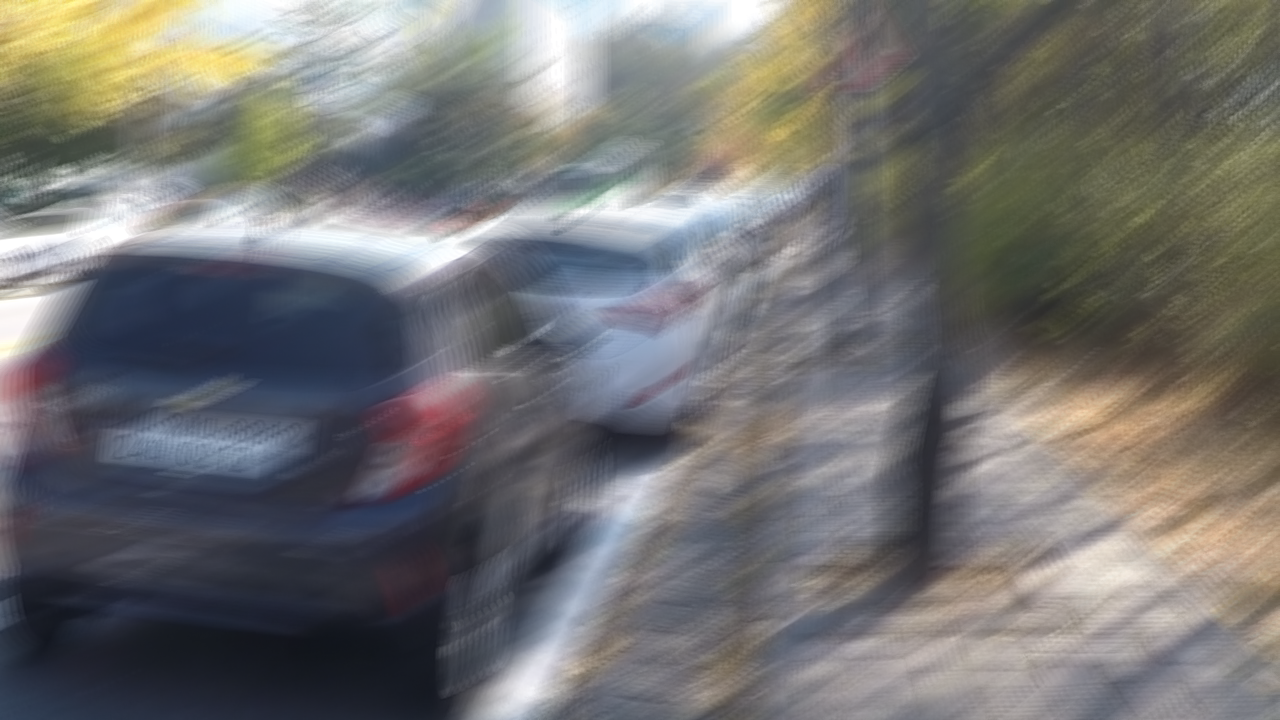}} &   
    \includegraphics[width=.13\textwidth,valign=t]{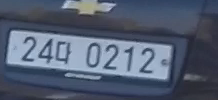}&
  	\includegraphics[width=.13\textwidth,valign=t]{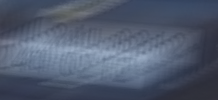}&   
      \includegraphics[width=.13\textwidth,valign=t]{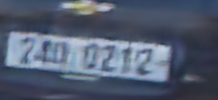}&
    \includegraphics[width=.13\textwidth,valign=t]{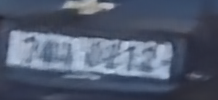}&
        \includegraphics[width=.13\textwidth,valign=t]{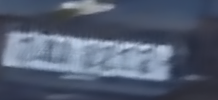}
  
\\
    &  \small~PSNR &\small~19.45 dB  & \small~23.85 dB & \small~23.56 dB & \small~23.86 dB   \\
    
    & \small~Reference & \small~Blurry    & \small~Gao \etal~\cite{gao2019dynamic}& \small~DBGAN~\cite{zhang2020dbgan}& \small~MTRNN~\cite{mtrnn2020} \\

    &

    \includegraphics[width=.13\textwidth,valign=t]{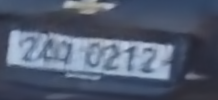}&  
     \includegraphics[width=.13\textwidth,valign=t]{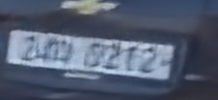}&
     \includegraphics[width=.13\textwidth,valign=t]{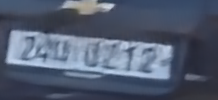}&
    \includegraphics[width=.13\textwidth,valign=t]{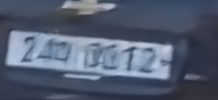}&
     \includegraphics[width=.13\textwidth,valign=t]{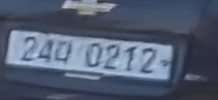}\\

     \small~19.45 dB & \small~24.85 dB 
     & \small~25.20 dB & \small 25.67 dB  & \small 24.33 dB & \small~\textbf{26.96 dB}\\
           \small~Blurry Image   & \small~DMPHN~\cite{dmphn2019} & \small~Suin \etal~\cite{Maitreya2020}   & \small MPRNet~\cite{Zamir_2021_CVPR_mprnet}& \small~MIMO-UNet+~\cite{cho2021rethinking_mimo} & \small~\textbf{\xnet }
\\
\end{tabular}}
\end{center}
\vspace{-6mm}
\caption{ \small  \underline{\textbf{Single image motion deblurring}} on GoPro~\cite{gopro2017}. \xnet generates sharper and visually-faithful result. 
}
\label{fig:deblurring}
\vspace{-0.5em}
\end{figure*}


\begin{table*}[!t]
\begin{center}
\caption{\underline{\textbf{Defocus deblurring}} comparisons on the DPDD testset~\cite{abdullah2020dpdd} (containing 37 indoor and 39 outdoor scenes). \textbf{S:} single-image defocus deblurring. \textbf{D:} dual-pixel defocus deblurring. \xnet sets new state-of-the-art for both single-image and dual pixel defocus deblurring. 
}
\label{table:dpdeblurring}
\vspace{-2mm}
\setlength{\tabcolsep}{10pt}
\scalebox{0.7}{
\begin{tabular}{l | c c c c | c c c c | c c c c }
\toprule[0.15em]
   & \multicolumn{4}{c|}{\textbf{Indoor Scenes}} & \multicolumn{4}{c|}{\textbf{Outdoor Scenes}} & \multicolumn{4}{c}{\textbf{Combined}} \\
\cline{2-13}
   \textbf{Method} & PSNR~$\textcolor{black}{\uparrow}$ & SSIM~$\textcolor{black}{\uparrow}$& MAE~$\textcolor{black}{\downarrow}$ & LPIPS~$\textcolor{black}{\downarrow}$  & PSNR~$\textcolor{black}{\uparrow}$ & SSIM~$\textcolor{black}{\uparrow}$& MAE~$\textcolor{black}{\downarrow}$ & LPIPS~$\textcolor{black}{\downarrow}$  & PSNR~$\textcolor{black}{\uparrow}$ & SSIM~$\textcolor{black}{\uparrow}$& MAE~$\textcolor{black}{\downarrow}$ & LPIPS~$\textcolor{black}{\downarrow}$   \\
\midrule[0.15em]
EBDB$_S$~\cite{karaali2017edge_EBDB} & 25.77 & 0.772 & 0.040 & 0.297 & 21.25 & 0.599 & 0.058 & 0.373 & 23.45 & 0.683 & 0.049 & 0.336 \\
DMENet$_S$~\cite{lee2019deep_dmenet}  & 25.50 & 0.788 & 0.038 & 0.298 & 21.43 & 0.644 & 0.063 & 0.397 & 23.41 & 0.714 & 0.051 & 0.349 \\
JNB$_S$~\cite{shi2015just_jnb} & 26.73 & 0.828 & 0.031 & 0.273 & 21.10 & 0.608 & 0.064 & 0.355 & 23.84 & 0.715 & 0.048 & 0.315 \\
DPDNet$_S$~\cite{abdullah2020dpdd} &26.54 & 0.816 & 0.031 & 0.239 & 22.25 & 0.682 & 0.056 & 0.313 & 24.34 & 0.747 & 0.044 & 0.277\\
KPAC$_S$~\cite{son2021single_kpac} & 27.97 & 0.852 & 0.026 & 0.182 & 22.62 & 0.701 & 0.053 & 0.269 & 25.22 & 0.774 & 0.040 & 0.227 \\
IFAN$_S$~\cite{Lee_2021_CVPRifan} & \underline{28.11}  & \underline{0.861}  & \underline{0.026} & \underline{0.179}  & \underline{22.76}  & \underline{0.720} & \underline{0.052}  & \underline{0.254}  & \underline{25.37} & \underline{0.789} & \underline{0.039} & \underline{0.217}\\
\textbf{Restormer}$_S$& \textbf{28.87}  & \textbf{0.882}  & \textbf{0.025} & \textbf{0.145} & \textbf{23.24}  & \textbf{0.743}  & \textbf{0.050} & \textbf{0.209}  & \textbf{25.98}  & \textbf{0.811}  & \textbf{0.038}  & \textbf{0.178}   \\
\midrule[0.1em]
\midrule[0.1em]
DPDNet$_D$~\cite{abdullah2020dpdd} & 27.48 & 0.849 & 0.029 & 0.189 & 22.90 & 0.726 & 0.052 & 0.255 & 25.13 & 0.786 & 0.041 & 0.223 \\
RDPD$_D$~\cite{abdullah2021rdpd} & 28.10 & 0.843 & 0.027 & 0.210 & 22.82 & 0.704 & 0.053 & 0.298 & 25.39 & 0.772 & 0.040 & 0.255 \\

Uformer$_D$~\cite{wang2021uformer} & 28.23 & 0.860 & 0.026 & 0.199 & 23.10 & 0.728 & 0.051 & 0.285 & 25.65 & 0.795 & 0.039 & 0.243 \\

IFAN$_D$~\cite{Lee_2021_CVPRifan} & \underline{28.66} & \underline{0.868} & \underline{0.025} & \underline{0.172} & \underline{23.46} & \underline{0.743} & \underline{0.049} & \underline{0.240} & \underline{25.99} & \underline{0.804} & \underline{0.037} & \underline{0.207} \\

\textbf{Restormer}$_D$& \textbf{29.48}  & \textbf{0.895}  & \textbf{0.023} & \textbf{0.134} & \textbf{23.97}  & \textbf{0.773}  & \textbf{0.047} & \textbf{0.175}  & \textbf{26.66}  & \textbf{0.833}  & \textbf{0.035}  & \textbf{0.155} \\
\bottomrule[0.1em]
\end{tabular}}
\end{center}
\vspace{-1.5em}
\end{table*}

\begin{figure*}[!t]
\begin{center}
\scalebox{0.99}{
\begin{tabular}[b]{c@{ } c@{ }  c@{ } c@{ } c@{ }   }\hspace{-4mm}
\multirow{4}{*}{\includegraphics[width=.36\textwidth,valign=t]{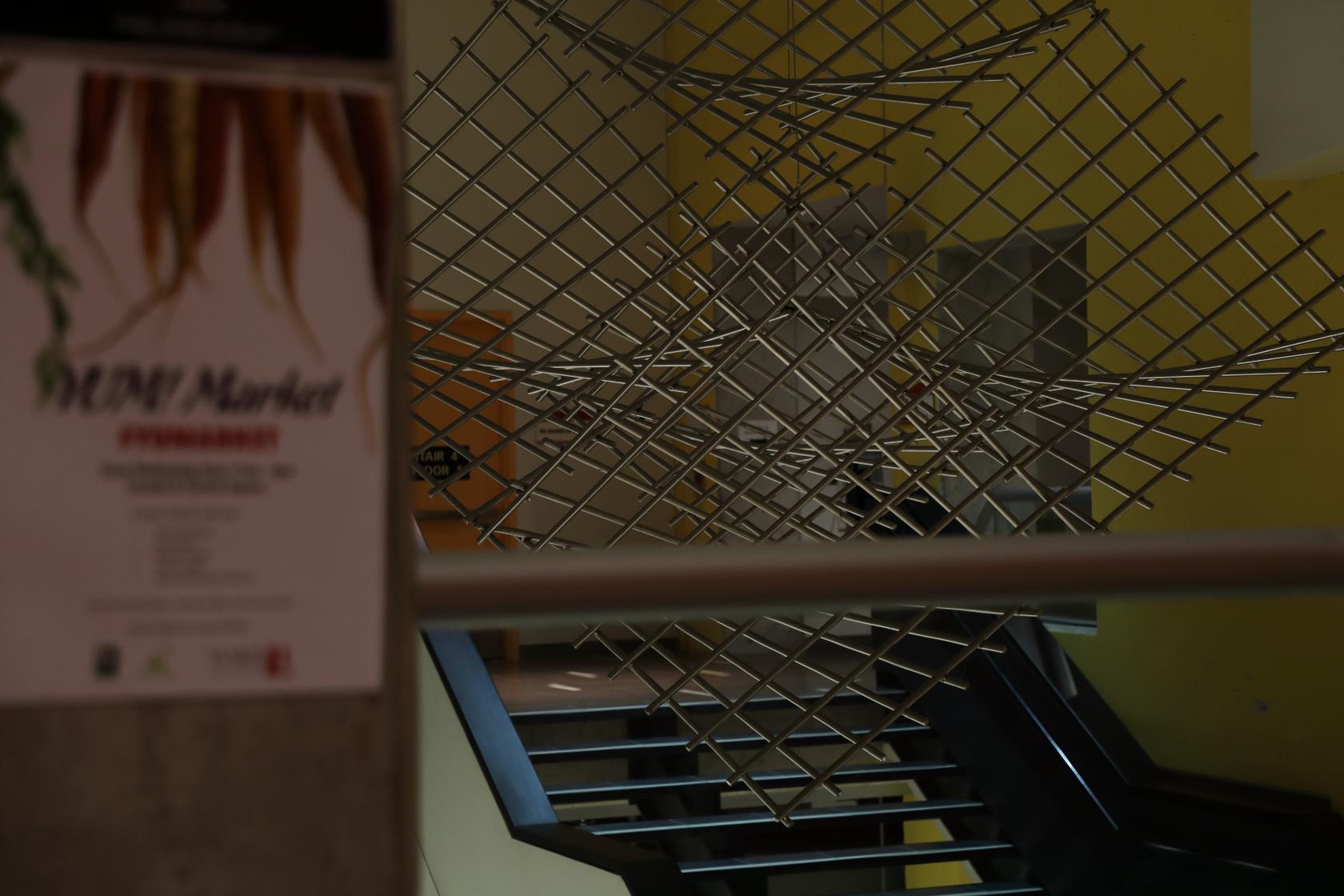}} &   
    \includegraphics[width=.145\textwidth,valign=t]{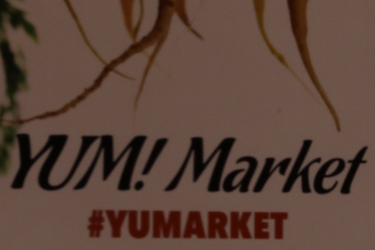} &
    \includegraphics[width=.145\textwidth,valign=t]{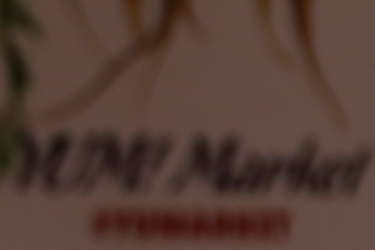} &
    \includegraphics[width=.145\textwidth,valign=t]{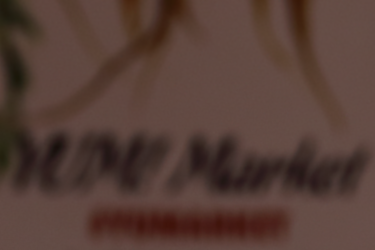} &
    \includegraphics[width=.145\textwidth,valign=t]{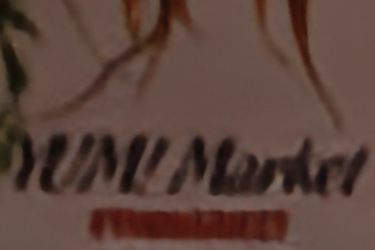}
\\
    &  \small~PSNR &\small~27.19 dB  & \small~27.44 dB & \small~28.67 dB   \\
    & \small~Reference & \small~Blurry  & \small~DMENet~\cite{lee2019deep_dmenet} & \small DPDNet~\cite{abdullah2020dpdd}  \\

    &
    \includegraphics[width=.145\textwidth,valign=t]{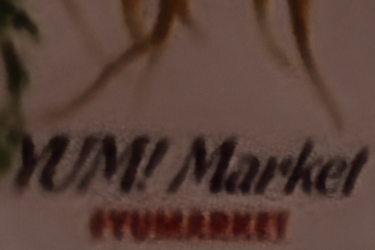} &
    \includegraphics[width=.145\textwidth,valign=t]{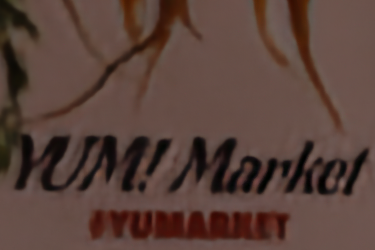} &
    \includegraphics[width=.145\textwidth,valign=t]{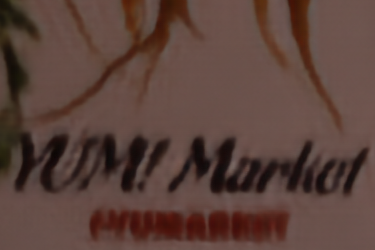} &
    \includegraphics[width=.145\textwidth,valign=t]{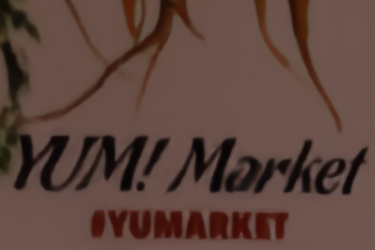}\\
     \small~27.19 dB & \small~29.01 dB & \small~28.35 dB & \small 29.12 dB & \small~\textbf{30.45 dB}\\
          \small~Blurry Image  & \small~RDPD~\cite{abdullah2021rdpd} & \small~IFAN~\cite{Lee_2021_CVPRifan} & \small Uformer\cite{wang2021uformer} &  \small~\textbf{\xnet}
\\
\end{tabular}}
\end{center}
\vspace*{-6mm}
\caption{\underline{\textbf{Dual-pixel defocus deblurring}} comparison on the DPDD dataset~\cite{abdullah2020dpdd}. Compared to the other approaches, our \xnet more effectively removes blur while preserving the fine image details.}
\label{fig:dualpixel deblurring}
\vspace{-1em}
\end{figure*}



\begin{table*}[t]
\parbox{.435\linewidth}{
\centering
\caption{\underline{\textbf{Gaussian grayscale image denoising}} comparisons for two categories of methods. Top super row: learning a single model to handle various noise levels. Bottom super row: training a separate model for each noise level.} 
\label{table:graydenoising}
\vspace{-2mm}
\setlength{\tabcolsep}{1.5pt}
\scalebox{0.7}{
\begin{tabular}{l | c c c | c c c | c c c}
\toprule[0.15em]
   & \multicolumn{3}{c|}{\textbf{Set12}~\cite{DnCNN}} & \multicolumn{3}{c|}{\textbf{BSD68}~\cite{martin2001database_bsd}} & \multicolumn{3}{c}{\textbf{Urban100}~\cite{huang2015single_urban100}} \\
 \cline{2-10}
   \textbf{Method} & $\sigma$$=$$15$ & $\sigma$$=$$25$ & $\sigma$$=$$50$ & $\sigma$$=$$15$ & $\sigma$$=$$25$ & $\sigma$$=$$50$ & $\sigma$$=$$15$ & $\sigma$$=$$25$ & $\sigma$$=$$50$ \\
\midrule[0.15em]
DnCNN~\cite{DnCNN}  &32.67 & 30.35 & 27.18 & 31.62 & 29.16 & 26.23 & 32.28 & 29.80 & 26.35\\
FFDNet~\cite{FFDNetPlus}  &32.75 & 30.43 & 27.32 & 31.63 & 29.19 & 26.29 & 32.40 & 29.90 & 26.50\\ 
IRCNN~\cite{zhang2017learning}  &32.76 & 30.37 & 27.12 & 31.63 & 29.15 & 26.19 & 32.46 & 29.80 & 26.22\\ 
DRUNet~\cite{zhang2021DPIR}  & \underline{33.25} & \underline{30.94} & \underline{27.90} & \underline{31.91} & \underline{29.48} & \underline{26.59} & \underline{33.44} & \underline{31.11} & \underline{27.96}\\ 
\textbf{\xnet} & \textbf{33.35}	& \textbf{31.04}	& \textbf{28.01} & \textbf{31.95}	& \textbf{29.51}	& \textbf{26.62} & \textbf{33.67}	& \textbf{31.39}	& \textbf{28.33}\\ 
\midrule[0.1em]
\midrule[0.1em]
FOCNet~\cite{jia2019focnet}  &33.07 & 30.73 & 27.68 & 31.83 & 29.38 & 26.50 & 33.15 & 30.64 & 27.40\\
MWCNN~\cite{liu2018MWCNN}  &33.15 & 30.79 & 27.74 & 31.86 & 29.41 & 26.53 & 33.17 & 30.66 & 27.42\\
NLRN~\cite{liu2018NLRN}  &33.16 & 30.80 & 27.64 & 31.88 & 29.41 & 26.47 & 33.45 & 30.94 & 27.49\\
RNAN~\cite{zhang2019residual}  &- & - & 27.70 & - & - & 26.48 & - & - & 27.65\\
DeamNet~\cite{ren2021adaptivedeamnet}  &33.19 & 30.81 & 27.74 & 31.91 & 29.44 & 26.54 & 33.37 & 30.85 & 27.53 \\
DAGL~\cite{mou2021dynamicDAGL}  &33.28 & 30.93 & 27.81 & 31.93 & 29.46 & 26.51 & 33.79 & 31.39 & 27.97 \\
SwinIR~\cite{liang2021swinir} & \underline{33.36} & \underline{31.01} & \underline{27.91} & \textbf{31.97} & \underline{29.50} & \underline{26.58} & \underline{33.70} & \underline{31.30} & \underline{27.98} \\ 
 \textbf{\xnet} & \textbf{33.42} & \textbf{31.08} & \textbf{28.00} & \underline{31.96} & \textbf{29.52}& \textbf{26.62}& \textbf{33.79} & \textbf{31.46}& \textbf{28.29}\\ 
\bottomrule[0.1em]
\end{tabular}}
}
\hfill
\parbox{.54\linewidth}{
\centering
\caption{\underline{\textbf{Gaussian color image denoising}}. Our \xnet demonstrates favorable performance among both categories of methods. On Urban dataset~\cite{huang2015single_urban100} for noise level $50$, \xnet yields $0.41$ dB gain over CNN-based DRUNet~\cite{zhang2021DPIR}, and $0.2$ dB over Transformer model SwinIR~\cite{liang2021swinir}.}
\label{table:colordenoising}
\vspace{-2mm}
\setlength{\tabcolsep}{1.5pt}
\scalebox{0.7}{
\begin{tabular}{l | c c c | c c c | c c c | c c c}
\toprule[0.15em]
   & \multicolumn{3}{c|}{\textbf{CBSD68}~\cite{martin2001database_bsd}} & \multicolumn{3}{c|}{\textbf{Kodak24}~\cite{kodak}} & \multicolumn{3}{c|}{\textbf{McMaster}~\cite{zhang2011color_mcmaster}} & \multicolumn{3}{c}{\textbf{Urban100}~\cite{huang2015single_urban100}} \\
 \cline{2-13}
   \textbf{Method} & $\sigma$$=$$15$ & $\sigma$$=$$25$ & $\sigma$$=$$50$ & $\sigma$$=$$15$ & $\sigma$$=$$25$ & $\sigma$$=$$50$ & $\sigma$$=$$15$ & $\sigma$$=$$25$ & $\sigma$$=$$50$ & $\sigma$$=$$15$ & $\sigma$$=$$25$ & $\sigma$$=$$50$ \\
\midrule[0.15em]
IRCNN~\cite{zhang2017learning}   & 33.86 & 31.16 & 27.86 & 34.69 & 32.18 & 28.93 & 34.58 & 32.18 & 28.91 & 33.78 & 31.20 & 27.70\\
FFDNet~\cite{FFDNetPlus}  &33.87 & 31.21 & 27.96 & 34.63 & 32.13 & 28.98 & 34.66 & 32.35 & 29.18 & 33.83 & 31.40 & 28.05 \\
DnCNN~\cite{DnCNN}  &33.90 & 31.24 & 27.95 & 34.60 & 32.14 & 28.95 & 33.45 & 31.52 & 28.62 & 32.98 & 30.81 & 27.59 \\
DSNet~\cite{peng2019dilated}  & 33.91 & 31.28 & 28.05 & 34.63 & 32.16 & 29.05 & 34.67 & 32.40 & 29.28 & - & - & -\\
DRUNet~\cite{zhang2021DPIR}  & \underline{34.30} & \underline{31.69} & \underline{28.51} & \underline{35.31} & \underline{32.89} & \underline{29.86} & \underline{35.40} & \underline{33.14} & \underline{30.08} & \underline{34.81} & \underline{32.60} & \underline{29.61} \\
\textbf{\xnet} & \textbf{34.39} & \textbf{31.78} & \textbf{28.59} & \textbf{35.44} & \textbf{33.02} & \textbf{30.00} & \textbf{35.55} & \textbf{33.31} & \textbf{30.29} & \textbf{35.06} & \textbf{32.91} & \textbf{30.02}\\ 
\midrule[0.1em]
\midrule[0.1em]
RPCNN~\cite{xia2020rpcnn}  &- & 31.24 & 28.06 & - & 32.34 & 29.25 & - & 32.33 & 29.33 & - & 31.81 & 28.62\\
BRDNet~\cite{tian2020BRDnet}  &34.10 & 31.43 & 28.16 & 34.88 & 32.41 & 29.22 & 35.08 & 32.75 & 29.52 & 34.42 & 31.99 & 28.56 \\
RNAN~\cite{zhang2019residual}  &-&-&28.27&-&-&29.58&-&-&29.72&-&-&29.08\\
RDN~\cite{zhang2020rdn}  &-&-&28.31&-&-&29.66&-&-&-&-&-&29.38\\
IPT~\cite{chen2021IPT}  &-&-&28.39&-&-&29.64&-&-&29.98&-&-&29.71\\
SwinIR~\cite{liang2021swinir} & \textbf{34.42} & \underline{31.78} & \underline{28.56} & \underline{35.34} & \underline{32.89} & \underline{29.79} & \underline{35.61} & \underline{33.20} & \underline{30.22} & \underline{35.13} & \underline{32.90} & \underline{29.82} \\
\textbf{\xnet} & \underline{34.40} & \textbf{31.79}& \textbf{28.60}& \textbf{35.47} & \textbf{33.04}& \textbf{30.01}& \textbf{35.61}& \textbf{33.34}& \textbf{30.30} & \textbf{35.13}& \textbf{32.96}& \textbf{30.02}\\ 
\bottomrule[0.1em]
\end{tabular}}
}
\vspace*{-1mm}
\end{table*}

\begin{table*}[t]
\begin{center}
\caption{\small \small \underline{\textbf{Real image denoising}} on SIDD~\cite{sidd} and DND~\cite{dnd} datasets. \textcolor{red}{$\ast$} denotes methods using additional training data. Our \xnet is trained only on the SIDD images and directly tested on DND. Among competing approaches, only \xnet surpasses $40$~dB PSNR.}
\label{table:realdenoising}
\vspace{-2mm}
\setlength{\tabcolsep}{2.5pt}
\scalebox{0.70}{
\begin{tabular}{c| c| c c c c c c c c c c c c c c c c }
\toprule[0.15em]
& \textbf{ Method} & DnCNN   & BM3D & CBDNet\textcolor{red}{*}   & RIDNet\textcolor{red}{*}  & AINDNet\textcolor{red}{*}  & VDN & SADNet\textcolor{red}{*} &DANet+\textcolor{red}{*} & CycleISP\textcolor{red}{*} & MIRNet & DeamNet\textcolor{red}{*} & MPRNet & DAGL &  Uformer &
 \textbf{\xnet} \\
\textbf{Dataset} & & \cite{DnCNN} & \cite{BM3D} & \cite{CBDNet} & \cite{RIDNet} & \cite{kim2020aindnet} & \cite{VDN} &  \cite{chang2020sadnet}	 & \cite{yue2020danet} &  \cite{zamir2020cycleisp} & \cite{zamir2020mirnet} & \cite{ren2021adaptivedeamnet}& \cite{Zamir_2021_CVPR_mprnet} & \cite{mou2021dynamicDAGL} & \cite{wang2021uformer} & (Ours) \\
\midrule[0.15em]
\textbf{SIDD} & PSNR~$\textcolor{black}{\uparrow}$ &  23.66 &  25.65 & 30.78  &  38.71  &  39.08  & 39.28  & 39.46  & 39.47 & 39.52  & 39.72  & 39.47  & 39.71 & 38.94 & \underline{39.77} & \textbf{40.02} \\
~\cite{sidd}  & SSIM~$\textcolor{black}{\uparrow}$ &  0.583 &  0.685 & 0.801  &  0.951  &  0.954  & 0.956  & 0.957  & 0.957 & 0.957  & 0.959  & 0.957  & 0.958 & 0.953 & \underline{0.959} & \textbf{0.960}\\
\midrule[0.1em]
\textbf{DND} & PSNR~$\textcolor{black}{\uparrow}$ &  32.43 &  34.51 & 38.06  &  39.26  &  39.37  & 39.38  & 39.59  & 39.58 & 39.56  & 39.88  & 39.63  & 39.80 & 39.77 & \underline{39.96} & \textbf{40.03} \\
~\cite{dnd}  & SSIM~$\textcolor{black}{\uparrow}$ &  0.790 &  0.851 & 0.942  &  0.953  &  0.951  & 0.952  & 0.952  & 0.955 & 0.956  & 0.956  & 0.953  & 0.954 & 0.956 & \underline{0.956} & \textbf{0.956} \\
\bottomrule
\end{tabular}}
\end{center}\vspace{-1.5em}
\end{table*}


\begin{figure*}[!t]
\begin{center}
\setlength{\tabcolsep}{1.5pt}
\scalebox{0.97}{
\begin{tabular}[b]{c c c c c c c c}
\includegraphics[width=.122\textwidth,valign=t]{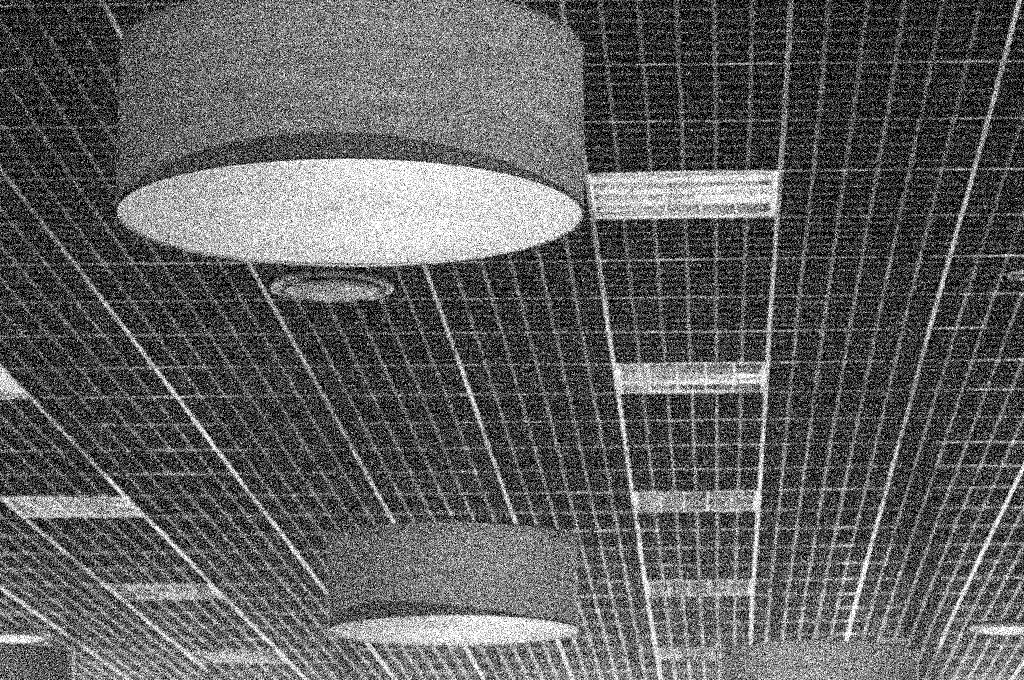} &   
\includegraphics[trim={ 0 34 80 20
 },clip,width=.122\textwidth,valign=t]{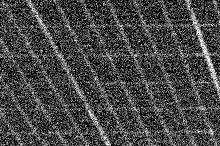} &   
\includegraphics[trim={ 0 34 80 20
 },clip,width=.122\textwidth,valign=t]{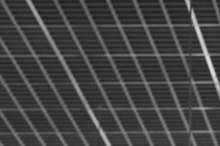} &   
\includegraphics[trim={ 0 34 80 20
 },clip,width=.122\textwidth,valign=t]{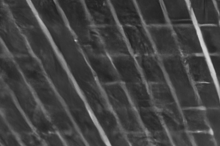} &   
\includegraphics[trim={ 0 34 80 20
 },clip,width=.122\textwidth,valign=t]{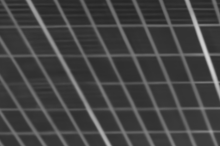} &   
\includegraphics[trim={ 0 34 80 20
 },clip,width=.122\textwidth,valign=t]{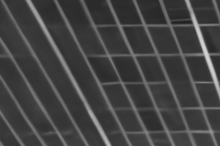} &   
\includegraphics[trim={ 0 34 80 20
 },clip,width=.122\textwidth,valign=t]{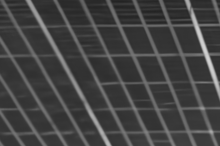} &   
\includegraphics[trim={ 0 34 80 20
 },clip,width=.122\textwidth,valign=t]{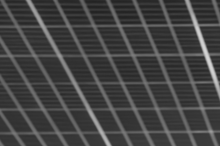} 
\\
\small~Noisy & \small~14.92 dB  & \small~PSNR & \small~27.61 dB &  \small~31.83 dB & \small~30.12 dB & \small~31.74 dB & \small~\textbf{32.83 dB}
\\
\small~Image & \small~Noisy & \small~Reference & \small~DnCNN~\cite{DnCNN} & \small~DRUNet~\cite{zhang2021DPIR} & \small~DeamNet~\cite{ren2021adaptivedeamnet}  & \small~SwinIR~\cite{liang2021swinir} & \small~\textbf{\xnet}
\\
\includegraphics[trim={ 0 0 0 0
 },clip,width=.122\textwidth,valign=t]{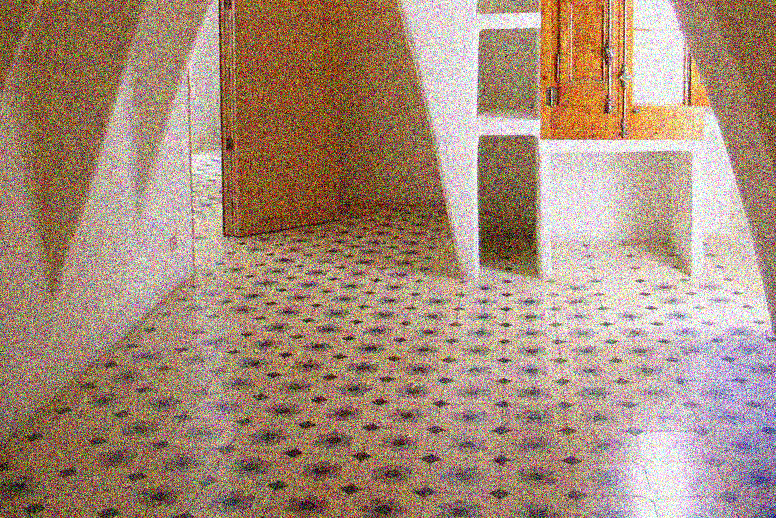} &   
\includegraphics[trim={ 80 0 20 40
 },clip,width=.122\textwidth,valign=t]{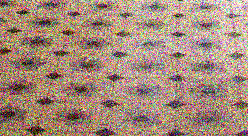} &   
\includegraphics[trim={ 80 0 20 40
 },clip,width=.122\textwidth,valign=t]{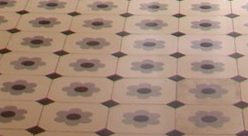} &   
\includegraphics[trim={ 80 0 20 40
 },clip,width=.122\textwidth,valign=t]{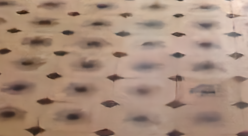} &   
\includegraphics[trim={ 80 0 20 40
 },clip,width=.122\textwidth,valign=t]{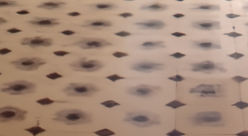} & 
 \includegraphics[trim={ 80 0 20 40
 },clip,width=.122\textwidth,valign=t]{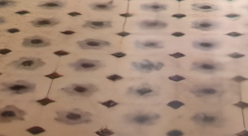} &  
\includegraphics[trim={ 80 0 20 40
 },clip,width=.122\textwidth,valign=t]{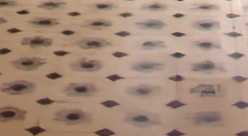} &   
\includegraphics[trim={ 80 0 20 40
 },clip,width=.122\textwidth,valign=t]{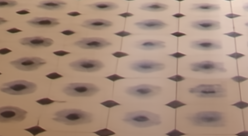} 
\\
\small~Noisy & \small~14.81 dB  & \small~PSNR & \small~33.83 dB &  \small~35.09 dB & \small~34.86 dB & \small~35.20 dB & \small~\textbf{35.63 dB}
\\
\small Image & \small Noisy & \small Reference & \small FFDNet~\cite{FFDNetPlus} &  \small DRUNet~\cite{zhang2021DPIR} & \small IPT~\cite{chen2021IPT} & \small SwinIR~\cite{liang2021swinir} & \small~\textbf{\xnet}
\\
\includegraphics[width=.122\textwidth,valign=t]{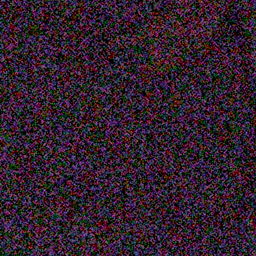} &   
\includegraphics[width=.122\textwidth,valign=t]{Images/RealDenoising/418input.png} &   
\includegraphics[width=.122\textwidth,valign=t]{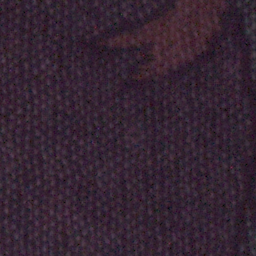} &   
\includegraphics[width=.122\textwidth,valign=t]{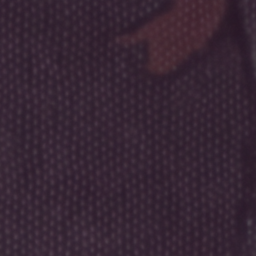} & 
\includegraphics[width=.122\textwidth,valign=t]{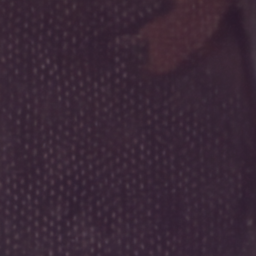} &  
\includegraphics[width=.122\textwidth,valign=t]{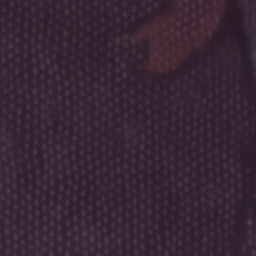} &   
\includegraphics[width=.122\textwidth,valign=t]{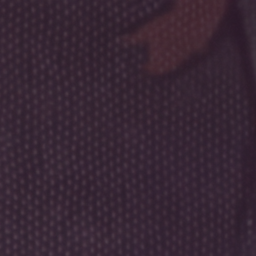} &   
\includegraphics[width=.122\textwidth,valign=t]{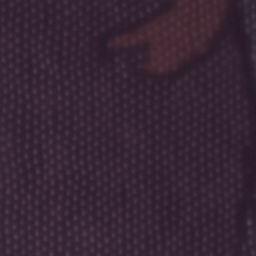} 
\\
\small~Noisy & \small~18.16 dB  & \small~PSNR & \small~31.36 dB & \small~30.25 dB  &  \small~31.17 dB & \small~31.15 dB & \small~\textbf{31.57 dB}
\\
\small Image & \small Noisy & \small Reference  & \small MIRNet~\cite{zamir2020mirnet} & \small DeamNet~\cite{ren2021adaptivedeamnet} & \small MPRNet~\cite{Zamir_2021_CVPR_mprnet} & \small Uformer~\cite{wang2021uformer} &  \small~\textbf{\xnet}
\end{tabular}}
\end{center}
\vspace*{-6mm}
\caption{\small Visual results on \underline{\textbf{Image denoising}}. Top row: Gaussian grayscale denoising. Middle row: Gaussian color denoising. Bottom row: real image denoising. The image reproduction quality of our \xnet is more faithful to the ground-truth than other methods. 
}
\label{fig:denoising}
\vspace{-0.5em}
\end{figure*}


\begin{table}[]
\begin{center}
\caption{\underline{\textbf{Ablation experiments for the Transformer block}}. PSNR is computed on a high-resolution Urban100 dataset~\cite{huang2015single_urban100}. }
\label{table:ablation main}
\vspace{-2mm}
\setlength{\tabcolsep}{1.3pt}
\scalebox{0.74}{
\begin{tabular}{c | l | c | c | c c c }
\toprule[0.15em]
 &  &  \textbf{FLOPs} &  \textbf{Params} & \multicolumn{3}{c}{\textbf{PSNR}} \\
\cline{5-7}
\textbf{Network} & \hspace{1cm} \textbf{Component} & (B) & (M) & $\sigma$=$15$ & $\sigma$=$25$ & $\sigma$=$50$ \\
\midrule[0.15em]
Baseline & \textbf{\textcolor{blue}{(a)}} UNet with Resblocks~\cite{lim2017enhanced_edsr} & 83.4 & 24.53 & 34.42 &	32.18 &	29.11 \\
\midrule[0.1em]

Multi-head & \textbf{\textcolor{blue}{(b)}} \hspace{0.6cm} MTA + FN~\cite{vaswani2017attention} & 83.7 & 24.84 & 34.66 & 32.39 & 29.28 \\
attention & \textbf{\textcolor{blue}{(c)}} \hspace{0.4cm} MDTA + FN~\cite{vaswani2017attention} & 85.3 & 25.02 & 34.72 & 32.48 & 29.43 \\

\midrule[0.1em]
Feed-forward & \textbf{\textcolor{blue}{(d)}} \hspace{0.55cm} MTA + GFN & 83.5 & 24.79 & 34.70 & 32.43 & 32.40 \\
network & \textbf{\textcolor{blue}{(e)}} \hspace{0.6cm} MTA + DFN & 85.8 & 25.08 & 34.68 & 32.45 & 29.42 \\
 & \textbf{\textcolor{blue}{(f)}} \hspace{0.65cm} MTA + GDFN & 86.2 & 25.12 & 34.77 & 32.56 & 29.54 \\
\midrule[0.1em]
Overall & \textbf{\textcolor{blue}{(g)}} \hspace{0.43cm} MDTA + GDFN & 87.7 & 25.31 & 34.82 & 32.61 & 29.62 \\
\bottomrule[0.1em]
\end{tabular}}
\end{center}
\vspace{-1.5em}
\end{table}


\subsection{Image Deraining Results}
We compute PSNR/SSIM scores using the Y channel in YCbCr color space in a way similar to 
existing methods\cite{mspfn2020,Zamir_2021_CVPR_mprnet,purohit2021spatially_spair}.
Table~\ref{table:deraining} shows that 
our \xnet achieves consistent and significant performance gains over existing approaches on all five datasets. 
Compared to the recent best method SPAIR~\cite{purohit2021spatially_spair}, \xnet achieves $1.05$~dB improvement when averaged across all datasets. On individual datasets, the gain can be as large as $2.06$~dB, \eg, Rain100L.
Figure~\ref{fig:deraining} shows a challenging visual example. Our \xnet reproduces a raindrop-free image while effectively preserving the structural content.

\subsection{Single-image Motion Deblurring Results}
We evaluate deblurring methods both on the synthetic datasets (GoPro~\cite{gopro2017}, HIDE~\cite{shen2019human}) and the real-world datasets (RealBlur-R~\cite{rim_2020_realblur}, RealBlur-J~\cite{rim_2020_realblur}).
Table~\ref{table:deblurring} shows that our \xnet outperforms other approaches on all four benchmark datasets. 
When averaged across all datasets, our method obtains {a} performance boost of $0.47$~dB over the recent algorithm MIMO-UNet+~\cite{cho2021rethinking_mimo} and $0.26$~dB over the previous best method MPRNet~\cite{Zamir_2021_CVPR_mprnet}.
Compared to MPRNet~\cite{Zamir_2021_CVPR_mprnet}, \xnet has $81\%$ fewer FLOPs (See \cref{Fig:teaser}). 
Moreover, our method shows $0.4$~dB improvement over the Transformer model IPT~\cite{chen2021IPT}, while having $4.4\times$ fewer parameters and runs $29\times$ faster. 
Notably, our \xnet is trained only on the GoPro~\cite{gopro2017} dataset, yet it demonstrates strong generalization to other datasets by setting new state-of-the-art. 
\cref{fig:deblurring} shows that the image produced by our method is more sharper and visually closer to the ground-truth than those of the other algorithms.

\subsection{Defocus Deblurring Results}
Table~\ref{table:dpdeblurring} shows image fidelity scores of the conventional defocus deblurring methods (EBDB~\cite{karaali2017edge_EBDB} and JNB~\cite{shi2015just_jnb}) as well as learning based approaches on the DPDD dataset~\cite{abdullah2020dpdd}. 
Our \xnet significantly outperforms the state-of-the-art schemes for the single-image and dual-pixel defocus deblurring tasks on all scene categories. 
Particularly on the combined scene category, \xnet yields ${\sim0.6}$~dB improvements over the previous best method IFAN~\cite{Lee_2021_CVPRifan}. 
Compared to the Transformer model Uformer~\cite{wang2021uformer}, our method provides a substantial gain of $1.01$~dB PSNR. Figure~\ref{fig:dualpixel deblurring} illustrates that our method is more effective in removing spatially varying defocus blur than other approaches. 

\subsection{Image Denoising Results}
We perform denoising experiments on synthetic benchmark datasets generated with additive white Gaussian noise (Set12~\cite{DnCNN}, BSD68~\cite{martin2001database_bsd}, Urban100~\cite{huang2015single_urban100}, Kodak24~\cite{kodak} and McMaster~\cite{zhang2011color_mcmaster}) as well as on real-world datasets (SIDD~\cite{sidd} and DND~\cite{dnd}). Following~\cite{mohan2019robust_biasfree,zhang2021DPIR,Zamir_2021_CVPR_mprnet}, we use bias-free \xnet for denoising.

\vspace{0.2em}
\noindent \textbf{Gaussian denoising.} 
Table~\ref{table:graydenoising} and Table~\ref{table:colordenoising} show PSNR scores of different approaches on several benchmark datasets for grayscale and color image denoising, respectively. 
Consistent with existing methods~\cite{zhang2021DPIR,liang2021swinir}, we include noise levels $15$, $25$ and $50$ in testing.
The evaluated methods are divided into two experimental categories: (1) learning a single model to handle various noise levels, and (2) learning a separate model for each noise level. Our \xnet achieves state-of-the-art performance under both experimental settings on different datasets and noise levels.
Specifically, for the challenging noise level $50$ on high-resolution Urban100 dataset~\cite{huang2015single_urban100}, \xnet achieves $0.37$~dB gain over the previous best CNN-based method DRUNet~\cite{zhang2021DPIR}, and $0.31$~dB boost over the recent transformer-based network SwinIR~\cite{liang2021swinir}, as shown in Table~\ref{table:graydenoising}.
Similar performance gains can be observed for the Gaussian color denoising in~Table~\ref{table:colordenoising}. 
It is worth mentioning that DRUNet~\cite{zhang2021DPIR} requires {the} noise level map as an additional input, whereas our method only takes the noisy image. Furthermore, compared to SwinIR~\cite{liang2021swinir}, our \xnet has $3.14\times$ fewer FLOPs and runs $13\times$ faster. Figure~\ref{fig:denoising} presents denoised results by different methods for grayscale denoising (top row) and color denoising (middle row). Our \xnet restores clean and crisp images.

\vspace{0.2em}
\noindent \textbf{Real image denoising.}
Table~\ref{table:realdenoising} shows that our method is the only one surpassing $40$~dB PSNR on both datasets.
Notably, on the SIDD dataset our \xnet obtains PSNR gains of $0.3$~dB and $0.25$~dB over the previous best CNN method MIRNet~\cite{zamir2020mirnet} and Transformer model Uformer~\cite{wang2021uformer}, respectively. \cref{fig:denoising} (bottom row) shows that our \xnet generates clean image without compromising fine texture.


\noindent%
\begin{table*}[bp]
\parbox{.27\textwidth}{
\centering
\captionof{table}{Influence of concat (w/o 1x1 conv) and refinement stage at \underline{\textbf{decoder (level-1)}}. We add the components to experiment Table~\ref{table:ablation main}\textcolor{blue}{\textbf{(g)}}. } 
\label{table:ablation concat}
\vspace{-2mm}
\setlength{\tabcolsep}{2pt}
\scalebox{0.7}{
\begin{tabular}{l | c | c | c}
\toprule[0.15em]
 & \textbf{PSNR} ($\sigma$=$50$) & \textbf{FLOPs} & \textbf{Params} \\
\midrule[0.15 em]
Table~\ref{table:ablation main}\textcolor{blue}{\textbf{(g)}} & 29.62 & 87.7 & 25.31 \\
\textbf{+} Concat & 29.66 & 110 & 25.65 \\
\textbf{+} Refinement & 29.71 & 141 & 26.12 \\
\bottomrule[0.1em]
\end{tabular}}
}
\hfill
\parbox{.35\textwidth}{
\centering
\captionof{table}{Results of training \xnet on fixed patch size and progressively large patch sizes. In \underline{\textbf{progressive learning}}~\cite{hoffer2019mix}, we reduce batch size (as patch size increases) to have similar time per optimization step as of fixed patch training. } 
\label{table:ablation progressive}
\vspace{-3mm}
\setlength{\tabcolsep}{2pt}
\scalebox{0.7}{
\begin{tabular}{l | c | c}
\toprule[0.15em]
\textbf{Patch Size} & \textbf{PSNR} ($\sigma$=$50$) & \textbf{Train Time (h)} \\
\midrule[0.15 em]
Fixed ($128^2$) & 29.71 & 22.5\\
Progressive ($128^2$ to $384^2$) & 29.78 & 23.1\\
\bottomrule[0.1em]
\end{tabular}}
}
\hfill
\parbox{.335\textwidth}{
\centering
\captionof{table}{\underline{\textbf{Deeper vs wider}} model. We adjust \# of transformer blocks to keep flops and params constant. Deep narrow model is more accurate, while wide shallow model is faster.} 
\label{table:ablation deepnarrow}
\vspace{-2mm}
\setlength{\tabcolsep}{4.5pt}
\scalebox{0.7}{
\begin{tabular}{l | c | c | c}
\toprule[0.15em]
\textbf{Dim} & \textbf{PSNR} ($\sigma$=$50$) & \textbf{Train Time (h)} & \textbf{Test Time (s)} \\
\midrule[0.15 em]
48 & \textbf{29.71} & 22.5 & 0.115\\
64 & 29.63 & 15.5 & 0.080\\
80 & 29.56 & \textbf{13.5} & \textbf{0.069}\\
\bottomrule[0.1em]
\end{tabular}}
}
\vspace*{-1em}
\end{table*}

\subsection{Ablation Studies}
For ablation experiments, we train Gaussian color denoising models on image patches of size $128$$\times$$128$ for $100$K iterations only. Testing is performed on Urban100~\cite{huang2015single_urban100}, and analysis is provided for a challenging noise level $\sigma{=}50$. FLOPs and inference time are computed on image size $256$$\times$$256$. Table~\ref{table:ablation main}-\ref{table:ablation deepnarrow} show that our contributions yield quality performance improvements. Next, we describe the influence of each component individually.

\noindent \textbf{Improvements in multi-head attention.} Table~\ref{table:ablation main}{\textcolor{blue}{c}} demonstrates that our MDTA provides favorable gain of $0.32$~dB over the baseline (Table~\ref{table:ablation main}{\textcolor{blue}{a}}). Furthermore, bringing locality to MDTA via depth-wise convolution improves robustness as removing it results in PSNR drop (see Table~\ref{table:ablation main}{\textcolor{blue}{b}}).

\noindent \textbf{Improvements in feed-forward network (FN).} Table~\ref{table:ablation main}{\textcolor{blue}{d}} shows that the gating mechanism in FN to control information flow yields $0.12$ dB gain over the conventional FN~\cite{vaswani2017attention}. As in multi-head attention, introducing local mechanism to FN also brings performance advantages (see  Table~\ref{table:ablation main}{\textcolor{blue}{e}}). We further strengthen the FN by incorporating gated depth-wise convolutions. Our GDFN (Table~\ref{table:ablation main}{\textcolor{blue}{f}}) achieves PSNR gain of $0.26$~dB over the standard FN~\cite{vaswani2017attention} for the noise level $50$. Overall, our Transformer block contributions lead to {a} significant gain of $0.51$~dB over the baseline.   

\noindent \textbf{Design choices for decoder at level-1.}
To aggregate encoder features with the decoder at level-1, we do not employ $1$$\times$$1$ convolution (that reduces channels by half) after concatenation operation. It is helpful in preserving fine textural details coming from the encoder, as shown in 
Table~\ref{table:ablation concat}.  These results further demonstrate the effectiveness of adding Transformer blocks in the refinement stage.

\noindent \textbf{Impact of progressive learning.} 
Table~\ref{table:ablation progressive} shows that the progressive learning provides better results than the fixed patch training, while having similar training time.   

\noindent \textbf{Deeper or wider \xnet?}
Table~\ref{table:ablation deepnarrow} shows that, under similar parameters/FLOPs budget, a deep-narrow model performs more accurately than its wide-shallow counterpart. However, the wider model runs faster due to parallelization. In this paper we use deep-narrow \xnet.

\section{Conclusion}
We present an image restoration Transformer model, \xnet, that is computationally efficient to handle high-resolution images. We introduce key designs to the core components of the Transformer block for improved feature aggregation and transformation. Specifically, our multi-Dconv head transposed attention (MDTA) module implicitly models global context by applying self-attention across channels rather than the spatial dimension, thus having linear complexity rather than quadratic. Furthermore, the proposed gated-Dconv feed-forward network (GDFN) introduces a gating mechanism to perform controlled feature transformation. 
To incorporate {the} strength of CNNs  into the Transformer model, both MDTA and GDFN modules include depth-wise convolutions for encoding spatially local context. Extensive experiments on 16 benchmark datasets demonstrate that \xnet achieves the state-of-the-art performance for numerous image restoration tasks.

\vspace{0.2em}
\noindent \textbf{Acknowledgements.} Ming-Hsuan Yang is supported by the NSF CAREER grant 1149783. Munawar Hayat is supported by the ARC DECRA Fellowship DE200101100. Special thanks to Abdullah Abuolaim and Zhendong Wang for providing the results. 

{\small
\bibliographystyle{ieee_fullname}
\bibliography{bib}
}

\end{document}